\documentclass{article}

\usepackage[preprint]{neurips_2026}

\usepackage[utf8]{inputenc} %
\usepackage[T1]{fontenc}    %
\usepackage{hyperref}       %
\usepackage{url}            %
\usepackage{booktabs}       %
\usepackage{amsfonts}       %
\usepackage{nicefrac}       %
\usepackage{microtype}      %
\usepackage{xcolor}         %

\usepackage{amsmath}
\usepackage{graphicx}
\usepackage{longtable}
\usepackage{caption}
\usepackage{makecell}
\usepackage{adjustbox}
\usepackage{tabularx}
\usepackage{makecell}
\usepackage{float}
\usepackage{pifont}
\newcommand{\xmark}{\ding{55}} 
\usepackage{printlen}
\usepackage{pdflscape} %
\usepackage{longtable} %
\usepackage{adjustbox}
\usepackage{fancyhdr}
\usepackage{tikz}

\fancypagestyle{lscape_footer}{
    \fancyhf{}
    
    \fancyfoot[C]{
        \begin{tikzpicture}[remember picture,overlay]
            \node[rotate=90, xshift=0cm, yshift=1.5cm] at (current page.east) {\thepage};
        \end{tikzpicture}
    }
}

\title{MAPS: A Synthetic Dataset for Probing Vision Models in a Controlled 3D Scene Space}

\author{%
Santiago Galella \\
  FIAS \& Institute of Computer Science \\
  Goethe University Frankfurt \\
  \texttt{galella@fias.uni-frankfurt.de} \\
  \And
  Pamela Osuna-Vargas \\
  FIAS \& Institute of Computer Science \\
  Goethe University Frankfurt \\
  \texttt{osuna@fias.uni-frankfurt.de} \\
  \And
  Maren Wehrheim \\
  Mila \& Department of Biology \\
  York University \\
  \texttt{maren.wehrheim@mila.quebec} \\
  \And
  Martina G. Vilas \\
   Institute of Computer Science \\
  Goethe University Frankfurt \\
  \texttt{martinagvilas@em.uni-frankfurt.de} \\
  \And
  Gemma Roig \\
   Institute of Computer Science \\
  Goethe University Frankfurt \\
  The Hessian Center for Artificial Intelligence  \\
  \texttt{roignoguera@em.uni-frankfurt.de} \\
  \And
  Matthias Kaschube\\
  FIAS \& Institute of Computer Science \\
  Goethe University Frankfurt \\
  \texttt{kaschube@fias.uni-frankfurt.de} \\
}

\begin{document}

\maketitle

\begin{abstract}
Modern vision models achieve strong performance on standard benchmarks, yet their aggregate accuracy reveals little about which scene properties drive their predictions. Existing robustness benchmarks provide important stress tests, but typically manipulate global 2D image properties, rely on entangled real-world variation, or cover only a limited set of 3D objects and scene parameters. We introduce MAPS (Manifolds of Artificial Parametric Scenes), a scalable instrument for controlled attribution of vision model behavior to scene parameters. MAPS comprises 2,618 curated photorealistic 3D meshes validated for recognizability across 560 ImageNet classes and provides a Blender-based rendering pipeline for on-demand image generation under continuous variation of nine independent scene-factors spanning background, camera, and lighting, extensible to other factors. To showcase its applicability, we use MAPS to evaluate 20 convolutional and transformer-based models by quantifying their reliance on these scene factors through regression-based sensitivity analysis. We find a near-universal failure axis across all tested architectures: camera distance and elevation consistently dominate recognition failure regardless of ImageNet accuracy. However, the full sensitivity structure reveals that modern CNNs and transformers cluster together, distinct from older architectures, suggesting that fine-grained architectural design choices, rather than the coarse CNN-versus-transformer distinction, are the stronger determinant of sensitivity profiles.

\end{abstract}

\section{Introduction}

Natural vision requires extracting stable object categories from scenes where viewpoint, illumination, and context vary continuously. Artificial vision systems trained for categorization, and analogously, biological systems performing recognition tasks, achieve their goal through representations that become increasingly invariant to category-irrelevant variation while preserving category-relevant structure \citep{Logothetis1996, Dicarlo2007, Dicarlo2012, Geirhos2020}. Characterizing how these representations depend on scene structure, i.e., which factors a model treats as a signal and which as a nuisance, helps understand what vision models have actually learned, and whether their decisions depend on regularities and biases of the training distribution rather than properties of the object categories themselves.

In natural images, scene factors such as lighting, pose, background, or object category are inherently entangled, making it difficult to isolate what drives model behavior. Vision models often exploit such correlations as shortcuts, achieving high accuracy by relying on contextual or spurious features rather than object-specific ones \citep{Ribeiro2016, Beery2018, Geirhos2020, Singla2021}. This limitation is especially relevant, as training data appears to be the dominant determinant of learned representations in vision models \citep{Hermann2020}. Moreover, architecture and training procedure can shape which shortcuts models acquire, allowing models with similar overall performance to rely on different visual cues, such as viewpoint, occlusion, shape, or texture \citep{Morrison2021, Naseer2021, Tuli2021} and to differ across many other quality dimensions beyond accuracy \citep{Hesse2025}. To disentangle these effects, a dataset is needed that captures how models depend on individual scene factors. %

Existing benchmarks provide only partial solutions. Natural-image robustness datasets inherit the entanglements they aim to probe and can manipulate only 2D image attributes such as corruption type or object placement \citep[e.g.,][]{Hendrycks2019, Li2023}. Synthetic 3D datasets enable independent variation of scene factors but are typically limited to simple shapes, cover few object classes, or expose only a narrow subset of scene parameters \citep[e.g.,][]{Dong2022, Johnson2017, Burgess2018, Gondal2019}. What is missing is a scalable, ImageNet-aligned evaluation framework in which a broad range of 3D scene factors can be varied independently and their effects on model behavior and internal representations can be quantified directly.

To fill this gap, we introduce MAPS (Manifolds of Artificial Parametric Scenes), a scalable, controllable, and ImageNet-aligned 3D benchmark with an open rendering pipeline to systematically evaluate scene-factor sensitivity in vision models. MAPS encompasses 2,618 manually curated photorealistic 3D meshes spanning 560 ImageNet classes, providing a substantially broader class coverage than prior controlled 3D benchmarks. We also provide a Blender-based API for on-demand image rendering under continuous variation of nine independent scene-factors: camera (azimuth, elevation, distance, and roll), light (azimuth, elevation, and power), and background (hue and saturation). MAPS is not intended as a fixed pre-rendered evaluation set; it exposes the parametric scene space itself, allowing users to design new sampling protocols, target specific factor combinations, or probe internal representations along smooth trajectories through the parameter space, supporting both behavioral and mechanistic interpretability research. To ensure dataset validity, we validate each mesh using CMA-ES \citep{Hansen2001} to confirm recognizability by pretrained vision models under at least one scene configuration, finding that all 2,618 meshes are correctly classified by at least one reference model under suitable scene parameters.

We use MAPS to evaluate 20 convolutional and transformer-based vision models, and quantify their reliance on each scene factor such as camera angle or background dependence through regression-based sensitivity analysis, thus providing a systematic benchmark for evaluating scene-factor sensitivity. This analysis reveals a systematic pattern in scene-factor reliance across models, with camera distance and elevation being the dominant drivers of failure across model architectures. Moreover, sensitivity profiles cluster models meaningfully, with modern CNNs and transformers forming a group distinct from older architectures, suggesting that fine-grained architectural design choices matter more than the coarse CNN-versus-transformer distinction. We also demonstrate that MAPS reveals differences in sensitivity profiles that are not captured by their overall categorization accuracy. Together, these results illustrate how MAPS reveals structured differences between model architectures and object categories that summary metrics like accuracy obscure.

Our main contributions are: 
\begin{itemize}
    \item a dataset with broad ImageNet-class coverage, comprising 2,618 curated and recognizability-validated photorealistic meshes from 560 ImageNet classes, exceeding prior controlled 3D benchmarks in both class diversity and mesh count;
    \item an on-demand rendering pipeline with continuous control over nine individual scene parameters and extensible to additional scene-factors and probing protocols;
    \item a scene-factor sensitivity framework that uses controlled rendering and regression to compare sensitivity profiles across models, revealing differences between architectures that are invisible to aggregate accuracy.
\end{itemize}

\section{Related Work}
\label{sec:related-work}

\paragraph{Synthetic datasets for evaluation of vision models.}

Since its introduction, ImageNet has been the standard benchmark for training and evaluating vision models, yet its reliance on natural images introduces unintended correlations between object identity, background, viewpoint, lighting, or scale, making it difficult to isolate the factors that drive model behavior. Most evaluation sets derived from ImageNet manipulate only global or weakly disentangled image properties, such as corruptions, background masking, attribute editing, or style transfer, rather than the underlying 3D scene structure \citep{Hendrycks2019, Hendrycks2021, Geirhos2019, Xiao2021, Li2023, Zhang2024}.  ObjectNet \citep{Barbu2019} combines natural images of objects with many viewpoint, rotation, and background variations, but offers only a small number of discrete conditions per factor and no mechanism to vary one factor while holding the others fixed. Synthetic datasets such as dSprites \citep{Matthey2017}, Shapes3D \citep{Burgess2018}, MPI-3D \citep{Gondal2019}, or CLEVR \citep{Johnson2017} address this by providing low-complexity controlled stimuli with independently manipulable factors. However, these datasets remain limited in visual realism and diversity of scene variations, making them less suitable for evaluating large-scale vision models trained on natural images.

Moving beyond simple 3D objects, several works have used 3D-rendered objects to study invariances and identifiability under more realistic conditions. Early datasets such as SmallNORB \citep{Lecun2004}, iLab-20M \citep{Borji2016}, or \citep{Pinto2011} provide photographs or rendered images under systematically varied lighting and viewpoints, but are limited to a small number of object categories. More recent datasets, including 3DIdent \citep{Zimmermann2021} and 3DIEBench \citep{Garrido2023} provide stronger factor control on small object collections, but are not aligned with ImageNet classes or designed to probe pretrained vision models. Large-scale collections of 3D objects such as ShapeNet \citep{Chang2015}, Objaverse-XL \citep{Deitke2023}, OmniObject3D \citep{Wu2023}, and S3D3C \citep{Spiess2024} have enabled progress in reconstruction, generation, and shape understanding. Although some have partial overlap with ImageNet classes \citep{Wu2023}, they are not curated for ImageNet-class coverage or equipped with a rendering pipeline to control scene parameters. Prior work \citep{Alcorn2019} demonstrates that DNNs misclassify the majority of the pose space for familiar ImageNet objects by using 3D-rendered scenes to systematically probe viewpoint sensitivity, but are limited to 30 objects from traffic-related classes. ViewFool \citep{Dong2022} renders 100 ImageNet-aligned objects with NeRFs to identify adversarial viewpoints, providing an important example of 3D-based probing of pretrained classifiers, but with limited category coverage and scene variability. ImageNet3D \citep{Ma2024} augments real ImageNet images with 3D pose annotations and provides 3D object models, but does not provide a controllable rendering pipeline. The closest prior work to ours is PUG:ImageNet \citep{Bordes2023}, which uses Unreal Engine to render 724 assets across 151 ImageNet classes under controlled variation of background, size, texture, orientation, and lighting. MAPS extends previous datasets along multiple axes, with increased class coverage and diversity (560 classes and 2,618 3D meshes), and higher flexibility, by offering an open source on-demand rendering pipeline instead of pre-rendered images or non-commercial applications.

\paragraph{Probing model representations.}

Computer vision models often exploit systematic regularities in their training data, including selection, capture, and negative-set biases \citep{Torralba2011}. Rather than learning robust object representations, models can rely on shortcuts that are predictive within a dataset but fail to generalize \citep{Geirhos2020, Ribeiro2016, Beery2018, Singla2021}. Such shortcuts include reliance on texture over shape \citep{Geirhos2019, Hermann2020}, local contour over global shape  \citep{Baker2018, Baker2020}, background and contextual cues vs. foreground information \citep{Zhu2016, Xiao2021}. Training data appears to be a major determinant of which regularities are learned \citep{Hermann2020}, and models with different architectures and training objectives can converge on their failure modes \citep{Geirhos2021}, while measurable differences in feature reliance exist across architecture families and training procedures \citep{Morrison2021, Naseer2021, Tuli2021}. Tools that can characterize these differences in a systematic fashion are thus necessary.

A range of methods has been developed to probe the learned representations in models. A standard method is linear probing \citep{Alain2016}, which evaluates whether information is decodable from intermediate model representations. Testing with Concept Activation Vectors (TCAV) \citep{Kim2018} extends this idea to human-interpretable concepts by training linear classifiers to separate activations from concept examples from those of random counter-examples. Regression Concept Vectors (RCVs) \citep{Graziani2018, Graziani2020} further generalizes this approach to continuous-valued concepts. Network Dissection \citep{Bau2017} instead aligns individual units with semantic concepts using supervised segmentation masks. A common limitation of these methods is that concepts are defined via curated image sets or annotated examples, in which the measured concept sensitivity can be confounded by uncontrolled correlations among scene properties such as background, object size or lighting. MAPS addresses this limitation by providing direct access to the ground-truth generative parameters of each image, making it possible to regress model behavior or intermediate activations against independently controlled scene factors, yielding interpretable sensitivity estimates based on direct manipulation of the scene rather than observation of curated examples.

\begin{table}[ht!]
    \caption{\textbf{Comparison of MAPS with related datasets for controlled evaluation of vision models.} \textasteriskcentered{}: rendering code is provided.}
    \label{tab:related-datasets}
    \centering
    \footnotesize
    \vspace{0.3cm} %
        \begin{tabular}{lccccccc}
            \toprule
                \textbf{Dataset} & \makecell{\textbf{Image} \\ \textbf{source}} & \makecell{\textbf{ImageNet-} \\ \textbf{aligned}} & \textbf{\# classes} & \makecell{\textbf{Controlled} \\ \textbf{factors}}
                & \makecell{\textbf{Indep.} \\ \textbf{control}}& \makecell{\textbf{On-} \\ \textbf{demand}} & \makecell{\textbf{Open} \\ \textbf{backend}}\\
            \midrule
                ImageNet-E \citep{Li2023}& 2D edited& \checkmark& 373& \makecell{Bg, dir.,\\size, pos.}& \checkmark & \xmark & \checkmark\\
                ObjectNet \citep{Barbu2019}& Photographs& \makecell{Partial (113)}& 313& \makecell{Cam, bg,\\pose (discrete)}& \makecell{\xmark\\(entangled)}& \xmark & n/a\\
                3DIEBench \citep{Garrido2023} & \makecell{3D\\rendered}& \xmark   & 55& \makecell{Cam, light,\\bg, pose} & \checkmark & \xmark (\textasteriskcentered{}) & \checkmark\\
                ViewFool \citep{Dong2022} & \makecell{3D\\rendered} & \checkmark   & 100 & Cam & \checkmark & \checkmark & \checkmark\\

                ImageNet3D \citep{Ma2024} & \makecell{Real + 3D\\rendered} & \checkmark   & 200 & \makecell{Pose\\(annot.)} & n/a & \xmark & n/a\\
                
                PUG:ImageNet \citep{Bordes2023} & \makecell{3D\\rendered} & \checkmark   & 151 & \makecell{Cam, bg, pose,\\texture,\\light, size} & \checkmark & \xmark & \xmark\\
            \midrule
                \textbf{MAPS} & \makecell{3D\\rendered} & \checkmark  & 560 & \makecell{Cam, bg,\\light} & \checkmark & \checkmark & \checkmark\\
                
            \bottomrule
        \end{tabular}
\end{table}

\section{Dataset}
\label{sec:dataset}

\begin{figure}[ht!]
  \centering
\begin{adjustbox}{max width=\textwidth}
  \includegraphics{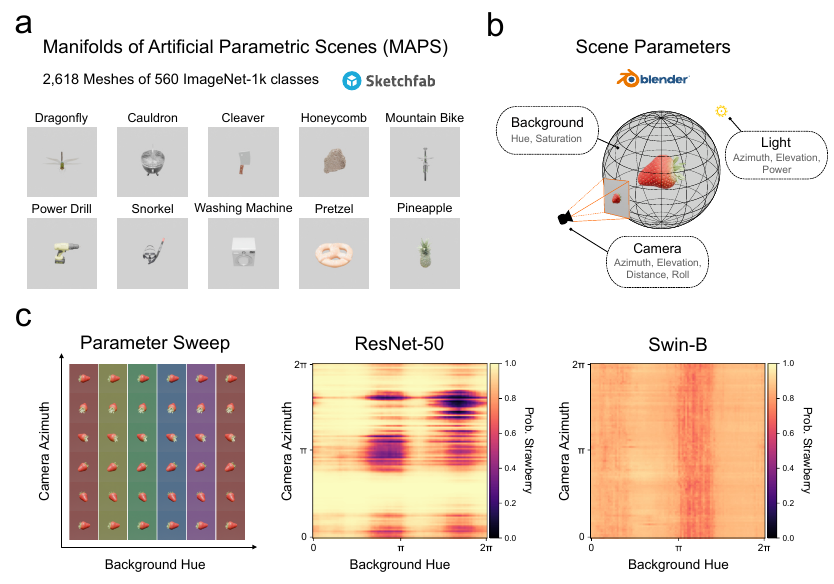}
  \end{adjustbox}
  \caption{\textbf{MAPS dataset.} (\textbf{a}) We collected 2,618 3D meshes from Sketchfab, encompassing 560 classes from ImageNet-1k. (\textbf{b})   \textbf{Scene parameters.} We additionally provide a flexible on-demand rendering pipeline implemented in Blender, to allow the rendering of 2D images with different scene parameters. (\textbf{c}) \textbf{Application.}  MAPS can be used to study representations in pretrained vision models. We find that model confidence is highly sensitive to scene parameters, with variations being model-specific (for all models refer to Fig.~\ref{figSI:confidence_models_2D}).}
\label{fig:fig1}
  \end{figure}

\paragraph{Mesh curation.}
To create the MAPS (Manifolds of Artificial Parametric Scenes) dataset, we manually curated a collection of high-quality 3D meshes from Sketchfab, a 3D-asset repository hosting millions of professional and community-contributed meshes. The final dataset contains 2,618 meshes spanning 560 unique ImageNet classes (Fig. \ref{fig:fig1}a), with at least one mesh per class and a median of 4 meshes per class. Mesh sources, author credits, and license information are accessible directly through the dataset. Full details of the curation process are provided in Appendix \ref{SI:mesh-curation}. %

\paragraph{Rendering pipeline.}
Our rendering pipeline imports each curated mesh into Blender 4.5.3 \citep{Blender2025} and renders it through a parametric pipeline that exposes nine independent scene factors (Fig. \ref{fig:fig1}b) grouped into three families: background (hue, saturation), camera (azimuth, elevation, distance, roll), and lighting (azimuth, elevation, power). These factors overlap with and extend the controllable parameters in previous datasets (Table~\ref{tab:related-datasets}) and capture common sources of real-world variation in object recognition while preserving experimental control. Each parameter is defined over a continuous value range, which enables reproducible sampling of the scene space (full ranges, definitions in Table \ref{tabSI:search_space}, with visualizations in Fig. \ref{figSI:goldfish-param-space}). %

Fig. \ref{fig:fig1}c shows how sweeping two parameters (background hue and camera azimuth) changes model confidence for a strawberry mesh, revealing differences between models. We show this analysis for all models in Fig. \ref{figSI:confidence_models_2D}. We additionally visualized internal representations, finding that intermediate layers recover the toroidal topology of a scene-parameter manifold in several architectures (Fig. \ref{figSI:activation-manifolds}). These results illustrate how MAPS can be used to probe the geometry and topology of internal representations along controlled scene trajectories. %

\begin{figure}[ht!]
  \centering
\begin{adjustbox}{max width=\textwidth}
  \includegraphics{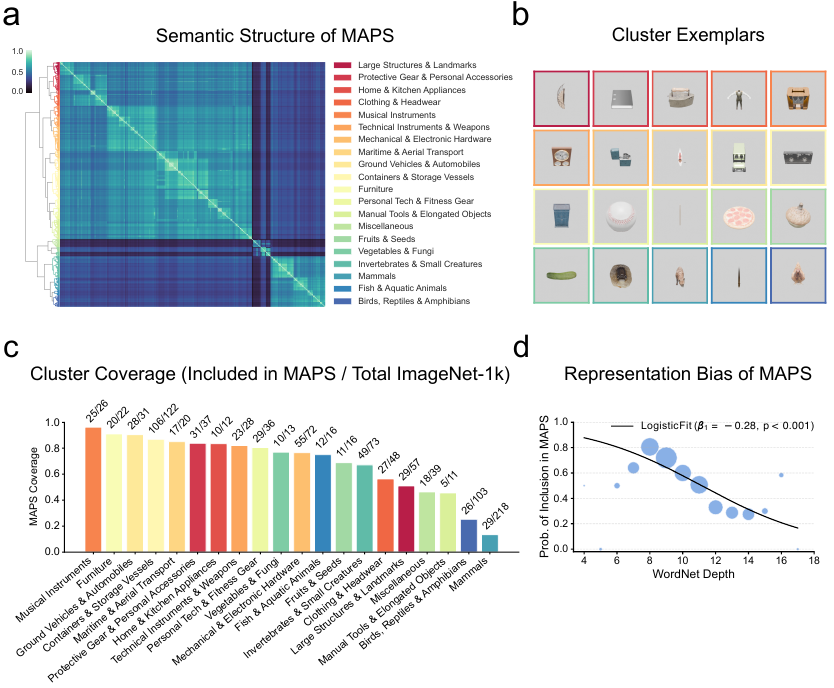}
  \end{adjustbox}
  \caption{\textbf{Estimating the semantic diversity of MAPS}.  \textbf{(a) Semantic structure.} Hierarchical clustering of the 560 MAPS classes based on pairwise  Wu–Palmer similarity between their WordNet synsets, using Ward's linkage. Cutting the dendrogram at 20 clusters yields semantically coherent groups, labeled by the lowest common WordNet hypernym of their cluster members. \textbf{(b) Representative cluster exemplars.} The medoid of each cluster, i.e., the class whose mean within-cluster similarity is highest, is shown with its rendered mesh. Border colors match the cluster colors in (a). \textbf{(c) Coverage of ImageNet-1k classes}. Per cluster, the fraction of ImageNet-1k classes under that cluster's WordNet root that are included in MAPS are shown as bars. Coverage is high for man-made subcategories (musical instruments, furniture, vehicles, $\geq80\%$) and low for mammals ($\sim 13$ \%), driven by ImageNet's 118 dog breeds, only 5 of which MAPS retains. \textbf{(d) Selection bias toward general categories.} Each dot shows the fraction of ImageNet-1k classes included in MAPS at a given WordNet depth; dot size indicates the number of ImageNet classes at that depth. The fitted logistic curve shows a negative association between WordNet depth and inclusion probability, indicating that deeper, more specific synsets are less frequently represented in MAPS. }
\label{fig:fig2}  
\end{figure}

\paragraph{Statistics and coverage.}
To characterize the semantic diversity of MAPS, we analyzed the WordNet structure of the included ImageNet classes. Specifically, we computed pairwise Wu-Palmer similarities \citep{Wu1994} between the WordNet synsets of all classes included in MAPS and applied hierarchical clustering using Ward's linkage \citep{Ward1963}, cutting the dendrogram at 20 clusters. The resulting groups span artifacts, tools, vehicles, instruments, animals, and household objects (Fig.~\ref{fig:fig2}a, b). We report more detailed clustering results in Table \ref{tab:full_cluster}. We find that coverage is highest for man-made objects, with musical instruments, furniture, vehicles, covering more than 80\% of the corresponding ImageNet classes (Fig.~\ref{fig:fig2}c) and lowest for mammals and birds, reflecting the fine-grained structure of ImageNet's animal subtree, which contains many dog breeds and bird species that are less aligned with ecologically typical object-level categorization \citep{Mehrer2021}. Finally, to quantify the relationship between taxonomic specificity and dataset inclusion, we fit a logistic function predicting MAPS inclusion from WordNet depth across all ImageNet-1k classes.  We find a significant negative association ($\beta_1=-0.28$, Wald $z=-8.87$, $p<0.001$; Fig.~\ref{fig:fig2}d), which corresponds to an approximately 24\% decrease in the odds of inclusion for each additional level of WordNet depth. This effect persists even after excluding the densely populated dog subtree ($\beta_1=-0.24$, Wald $z=-7.73$, $p<0.001$), confirming that the observed representation bias is a global characteristic of the dataset. These results demonstrate that while MAPS provides broad semantic coverage, it naturally prioritizes prototypical categories over highly fine-grained synsets due to 3D asset availability.

\paragraph{Mesh recognizability validation.}

Although all meshes in MAPS were visually inspected to ensure they are representative of their assigned ImageNet classes, visual inspection alone does not guarantee that the rendered objects are recognizable to standard vision models. Validation is essential as failures in downstream sensitivity analyses could otherwise reflect invalid or ambiguous geometry rather than genuine sensitivity to scene-factor variation. We therefore test whether each mesh can be correctly classified by a pretrained reference model under at least one scene configuration.

To do so, we use Covariance Matrix Adaptation Evolution Strategy (CMA-ES) \citep{Hansen2001} to search the nine-dimensional scene-parameter space (Fig.~\ref{figSI:optimization-pipeline}). Because the rendering pipeline is non-differentiable, gradient-free optimization is required. Here we use CMA-ES to find any combination of parameters that yields a correct top-1 classification. For each mesh, CMA-ES samples a population of candidate scene configurations from its current search distribution, renders each as an image, and ranks the population by the log-probability of the target class under three models, AlexNet, ResNet-50 and ViT-B/16. The mean and covariance of the search distribution are then adjusted to higher-scoring regions of the parameter space, and the next generation is drawn from the updated distribution. Optimization stops when the model assigns the highest probability to the correct ImageNet class for at least one candidate image, or when the maximum number of generations is reached. Implementation details and search ranges can be found in the Appendix ~\ref{SI:cmaes}. %

Under the full scene-parameter space, AlexNet correctly classified 2,581/2,618 meshes (98.6\%), ResNet-50 2,604/2,618 (99.5\%), and ViT-B/16 2,605/2,618 (99.5\%, Fig.~\ref{figSI:cmaes-statistics}). Five meshes failed recognition across all three models. Visual inspection revealed these to be mostly elongated objects (bullet train, chain, and steam locomotive) that occupy very few pixels at the minimum camera distance of 1.0 when placed at the scene origin (Fig.~\ref{figSI:cmaes-failure-models}). Constraining the camera distance to $[0,1]$ (instead of $[1, 8]$, Table~\ref{tabSI:search_space}) yielded correct classification by ViT-B/16 for all five, confirming that failure reflected camera placement geometry rather than unrecognizable object structure. Notably, most objects are classified correctly with only a small number of CMA-ES generations (median: 7 generations for ViT-B/16) and most require only a single optimization run (82.5\%; 2,159/2,618 for ViT-B/16). All 2,618 meshes are retained in the full MAPS database. Overall, this validation indicates that the curated meshes are recognizable under suitable scene configurations and that valid renderings are typically easy to find.

\section{Evaluations}
\label{sec:evaluation}

\paragraph{Setup.}

We use MAPS to characterize scene-factor sensitivity across pretrained vision models. Specifically, we ask which scene-factors most strongly influence model behavior, and whether these sensitivity profiles differ across architectures. To this end, we evaluate 20 models spanning convolutional (AlexNet, ConvNeXt-Tiny/Small/Base, DenseNet-121/161/201, EfficientNet-B0, GoogLeNet, Inception-v3, ResNet-18/50/101, VGG-16/19) and transformer-based architectures (ViT-B/16, ViT-L/16, Swin-B/S/T). The full list is in Table~\ref{tab:breakdown-probed-models}. We chose only models from TorchVision \citep{Torchvision2016} that are widely used in computer vision, were pretrained on ImageNet-1k for a fair comparison between models, and span a range of ImageNet-1k performance levels.

To keep the evaluation tractable while preserving semantic diversity, we select 100 ImageNet classes from MAPS by taking the class representing the medoid of each of the 20 WordNet-derived semantic clusters (Fig.~\ref{fig:fig2}a,b) together with its four nearest neighbors in semantic space. For each selected class, we include all mesh instances that passed the recognizability validation described in Section \ref{sec:dataset}.

\begin{figure}
  \centering
\begin{adjustbox}{max width=\textwidth}
  \includegraphics{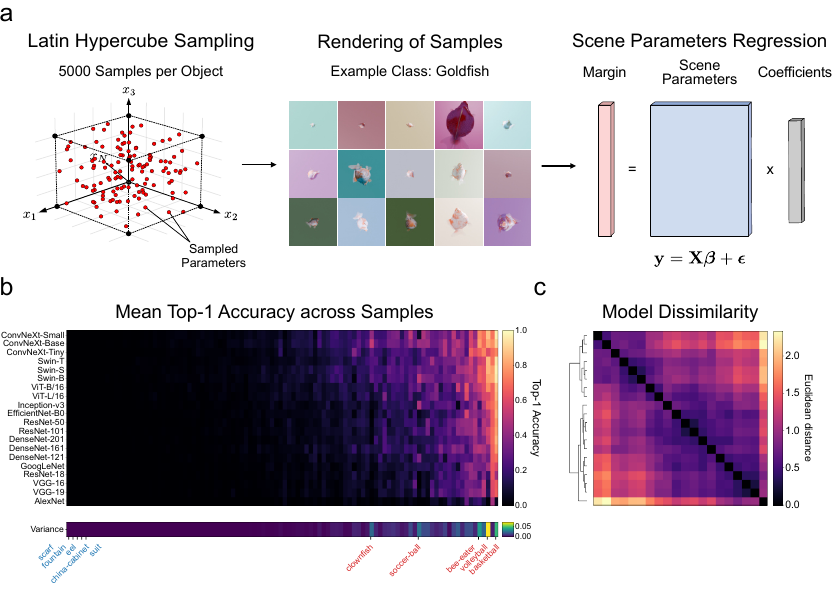}
  \end{adjustbox}
  \caption{\textbf{Scene-factor sensitivity analysis.}
  \textbf{(a) Methodology pipeline.} For each mesh, 5,000 scene configurations are drawn via Latin hypercube sampling over the nine-dimensional scene-parameter space (left). Each configuration is rendered through the MAPS API (center). For each model-mesh pair, we fit a regression predicting the decision's margin from the scene parameters (right).  \textbf{(b) Average top-1 accuracy heatmap across models (rows) and the 100 evaluated classes (columns).} Models were clustered and rows sorted. The bottom strip shows the per-class accuracy variance across models. Labels are displayed for the classes with the top-5 highest (red) and lowest (blue) accuracy variance across models. See Fig.~\ref{figSI:max-acc-lhs-maps} for maximum top-1 accuracy. \textbf{(c) Model dissimilarity.} Sorted dissimilarity matrix based on clustering of the rows in (b).}
  \label{fig:fig3}
\end{figure}

\paragraph{Sensitivity analysis protocol.}
For each mesh, we render 5,000 images by sampling the nine-dimensional scene-parameter space (camera azimuth, elevation, distance, roll; light azimuth, elevation, power; background hue, saturation) using Latin hypercube sampling (LHS) \citep{Mckay1979} (see Table~\ref{tabSI:search_space} for the search space range). LHS provides a broad coverage of the joint parameter space at a cost far below uniform grid sampling (Fig.~\ref{fig:fig3}a). Angular parameters are represented by their sine and cosine components to preserve cyclic structure under linear regression. For each model-mesh pair, we fit two surrogate regression models, a linear and a polynomial,  that relate scene parameters to the model's decision margin for the correct class and the largest logit among the remaining classes: \[y_c(\boldsymbol{\theta}) = z_c(\boldsymbol{\theta}) - \max_{i \neq c} z_i (\boldsymbol{\theta}),\] where $z_i$ denotes the pre-softmax logit for class $i$ and $\boldsymbol{\theta}$ is the scene-parameter vector. We regress on this margin rather than correct-class logits since the margin (i) directly indicates whether the model classifies correctly through its sign, and (ii) controls for scene-parameter effects that shift all logits without a change in the model's decision. The linear model captures first-order effects, and the polynomial model adds quadratic and pairwise interaction terms to capture non-linear effects. Both predictors (scene parameters) and target are z-scored within each model-mesh pair across the 5,000 LHS samples. Reported coefficient values are thus comparable across meshes and across models.

\paragraph{Model accuracy under controlled scene variation.}
We first quantify how each model performs on average across the LHS-sampled scene configurations. This analysis reveals substantial variation in top-1 accuracy across both models and classes (Fig.~\ref{fig:fig3}b). Older convolutional architectures (AlexNet, VGG-16) generally show lower accuracy under scene-factor variation, whereas transformer-based models maintain higher performance across a broader range of classes. At the class level, some classes are consistently recognized across models and scene configurations, such as soccer ball, %
whereas others are difficult for nearly all models, such as eel. These results show that MAPS exposes both model-specific robustness differences and class-specific failure modes that would be difficult to isolate in natural-image benchmarks. Fig.~\ref{figSI:max-acc-lhs-maps} shows for each model which objects are correctly classified at least once out of the 5000 sampled images.

\begin{figure}[t!]
  \centering
\begin{adjustbox}{max width=\textwidth}
  \includegraphics{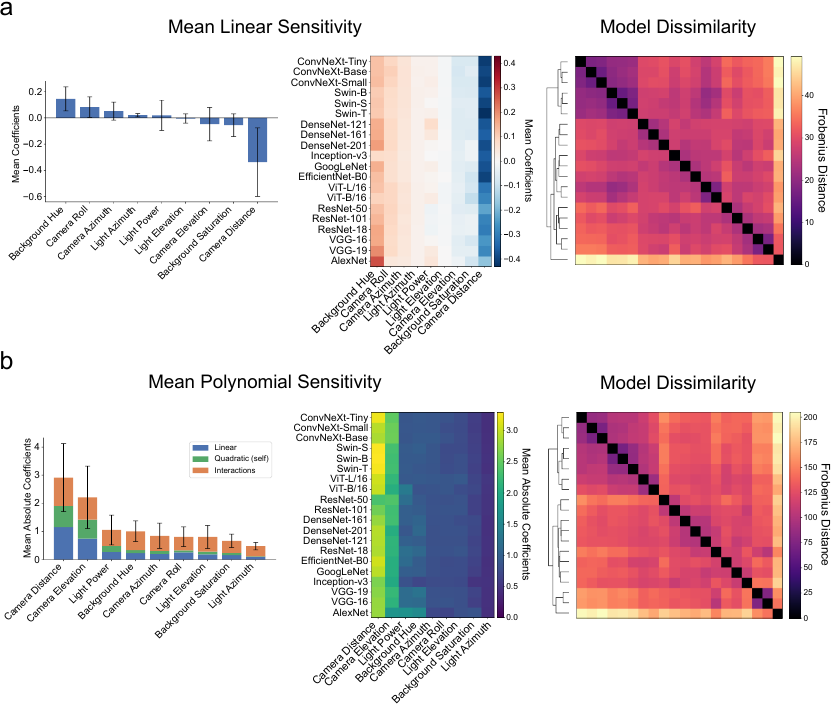}
 \end{adjustbox}
 \caption{\textbf{Scene-factor sensitivity for linear and polynomial regression.} For each model-mesh pair, we fit two regression models predicting the decision margin from the nine scene parameters: (a) a linear model, and (b) a polynomial model with linear, quadratic (self), and pairwise interaction terms. Coefficients are averaged across all the evaluated meshes, obtaining one coefficient profile per model. \textbf{(a, left)} Mean linear coefficient per scene parameter, averaged across all models, with their standard deviation displayed as error bars. \textbf{(a, middle)} Per-model linear coefficient profiles, averaged across meshes. Each row is a model, and each column a scene parameter. Rows are ordered by hierarchical clustering shown on the right (dendrogram). \textbf{(a, right)} Pairwise Frobenius distance between model coefficient matrices (objects × scene parameters, linear coefficients). \textbf{(b, left)} Mean absolute polynomial coefficients per scene parameter averaged across models, and decomposed into their linear (blue), quadratic (green), and pairwise interaction (orange) contributions. Error bars display the standard deviation across models. \textbf{(b, middle, right)} As for (a) but for the results on polynomial regression.}
\label{fig:regression-results}
\end{figure}

\paragraph{Scene-parameter sensitivity analysis.}
Regressing scene-parameters onto the model's decision margins using linear regression identifies camera-related factors as the dominant predictors across the tested models (Fig.~\ref{fig:regression-results}a, left). Camera distance has the largest mean absolute coefficient across models and consistently shows a negative association with model performance, indicating that increased camera-to-object distance reduces the margin between the correct class and its strongest competitor, pushing the model towards misclassification (Fig.~\ref{fig:regression-results}a, left). Additionally, camera roll, background saturation, and camera elevation have considerable impact on the margin, while lighting parameters (power, elevation, azimuth) contribute less. The polynomial regression reveals a structured non-linear effect: camera elevation has a strong quadratic effect (Fig.~\ref{fig:regression-results}b, left), indicating that decision margin deteriorates disproportionately at extreme low and high viewing angles relative to a more frontal viewpoint. These findings illustrate the use of MAPS for controlled scene-factor analysis: because scene parameters are explicitly controlled, model failures can be attributed to specific visual factors rather than to uncontrolled correlations in natural images. Fig. \ref{figSI:r_squared_fits} shows the explained variance for the linear and polynomial fits across objects and models. Additional Figs.~\ref{figSI:r2-convergence-linear} to ~\ref{figSI:stability-last-poly} show the convergence of the different scene parameters in the linear and polynomial regression for different sampling sizes.

\paragraph{Cross-model differences in sensitivity profiles.}
To compare sensitivity profiles across models, we first compute the average sensitivity profile across objects (Fig.~\ref{fig:regression-results}a, b, center panels). We then calculate the pairwise Frobenius norm distances between the models' coefficient matrices (objects × scene parameters), followed by hierarchical clustering to visualize model dissimilarity (Fig.~\ref{fig:regression-results}a, b, right panels). The resulting clustering suggests that models share a broad common structure, as the rank ordering of dominant scene-parameter effects is broadly consistent across architectures, with camera distance and elevation among the top three predictors in all models, independent of their ImageNet top-1 accuracy. Secondly, finer-grained differences in the full coefficient profile show architecture-specific patterns. While AlexNet is a clear outlier in both linear and polynomial analyses, more recent CNNs and transformer-based models form distinct clusters. This analysis shows that MAPS captures differences between models that are not apparent from aggregate accuracy alone. For example, ResNet-18 and GoogLeNet are matched on both ImageNet top-1 accuracy (Table~\ref{tab:breakdown-probed-models} and aggregate MAPS-LHS accuracy, Fig.~\ref{figSI:acc-maps-vs-inet}; see Fig.~\ref{figSI:class-acc-maps-vs-inet} for per-class accuracy), yet differ markedly in their sensitivity to background saturation and light power, as reflected in their mean coefficients, and this divergence in their sensitivity is directly visible at the mesh level: when a strawberry mesh is rendered across joint variation of two parameters, the two models display clearly different prediction-confidence patterns (Fig. \ref{figSI:confidence_models_2D}). %

\section{Discussion}
\label{sec:limitations-future-work}

MAPS provides a programmable, scalable, and ImageNet-aligned framework for probing how vision models respond to controlled changes in 3D scene structure. By pairing curated photorealistic meshes with on-demand rendering, MAPS enables fine-grained sensitivity analyses that natural-image benchmarks cannot directly support.

MAPS is designed to compare model sensitivity to scene factor variation in a reproducible reference frame, and is not intended to measure absolute robustness in deployment. Although the rendering pipeline currently exposes camera, lighting, and background parameters, manipulations of contextual objects, ground-geometry, occlusions, or texture are not yet offered. Also, class coverage reflects some selection biases: MAPS includes more inanimate objects than animals and prioritizes prototypical classes over fine-grained synsets, partly because ImageNet overrepresents fine-grained animal categories such as dog breeds.

Our evaluations reveal two complementary patterns that overall accuracy does not capture. First, camera distance and elevation emerge as dominant drivers of recognition failure across all evaluated models, directly quantifying a capture bias \citep{Torralba2011} in ImageNet's training distribution: models fail systematically at the camera distances and elevations that are presumably underrepresented during training, a failure mode shared across architectures rather than driven by specific architectural choices. Second, beyond this shared failure axis, models that match in accuracy can differ in their sensitivity profiles, thus reaching their decisions through different scene-factor reliance patterns. These findings extend a growing literature on architectural and training-recipe effects for feature reliance \citep{Morrison2021, Naseer2021, Tuli2021} by providing the parametric scene space as a common reference frame to measure such effects directly rather than inferring them from accuracy gaps on natural-image benchmarks. Notably, modern CNNs such as ConvNeXt \citep{Liu2022} cluster with transformer-based models rather than with older CNNs. This convergence is consistent with the architectural choices introduced in ConvNeXt, including patchify stems, large depthwise kernels, and LayerNorm, which were explicitly designed to close the gap with transformer architectures. This suggests that the CNN-versus-transformer distinction may be better understood as a continuum of architectural design choices, where models sharing key design decisions exhibit similar scene-factor sensitivity profiles regardless of their nominal family.%

The sensitivity analysis reported here is one of several use cases supported by MAPS. Because the generative parameters of each rendered image are exposed, the same controlled stimuli can be used to study internal representations, characterizing how scene structure is encoded across layers and across architectures, as well as its evolution over the training process. The dataset also defines a parametric scene space whose structure can support geometrical and topological analyses of representational manifolds along controlled scene trajectories in both artificial and biological systems \citep{Lin2024}.

In future work, differentiable rendering pipelines \citep{Jakob2022, Beilharz2025} could enable gradient-based search through the scene space, supporting a more efficient discovery of failure modes. More broadly, by combining broad class coverage with controlled stimuli and an open rendering pipeline, MAPS positions itself as a shared reference frame for the comparison of vision model behavior and with further potential to compare representational spaces in artificial and biological vision systems. %

\section{Acknowledgements}

We would like to thank Robin Hesse and all members of the ARENA unit for helpful discussions. This research was supported by the German Research Foundation (DFG) - DFG Research Unit FOR 5368 ARENA. MW received funding from the Connected Minds Postdoctoral Fellowship (supported by CFREF), and the Deutsche Forschungsgemeinschaft (DFG, German Research Foundation) – 414985841.

\bibliographystyle{plain}
\bibliography{bibliography}

@article{Ward1963,
  title={Hierarchical grouping to optimize an objective function},
  author={Ward Jr, Joe H},
  journal={Journal of the American statistical association},
  volume={58},
  number={301},
  pages={236--244},
  year={1963},
  publisher={Taylor \& Francis}
}

@article{Mckay1979,
  title={A Comparison of Three Methods for Selecting Values of Input Variables in the Analysis of Output from a Computer Code},
  author={McKay, MD and Beckman, RJ and Conover, WJ},
  journal={Technometrics},
  pages={239--245},
  year={1979},
  publisher={JSTOR}
}

@inproceedings{Wu1994,
  title={Verb semantics and lexical selection},
  author={Wu, Zhibiao and Palmer, Martha},
  booktitle={32nd annual meeting of the association for computational linguistics},
  pages={133--138},
  year={1994}
}

@article{Logothetis1996,
  title={Visual object recognition.},
  author={Logothetis, Nikos K and Sheinberg, David L},
  journal={Annual review of neuroscience},
  volume={19},
  pages={577--621},
  year={1996}
}

@article{Hansen2001,
  title={Completely derandomized self-adaptation in evolution strategies},
  author={Hansen, Nikolaus and Ostermeier, Andreas},
  journal={Evolutionary computation},
  volume={9},
  number={2},
  pages={159--195},
  year={2001},
  publisher={MIT Press}
}

@inproceedings{Lecun2004,
  title={Learning methods for generic object recognition with invariance to pose and lighting},
  author={LeCun, Yann and Huang, Fu Jie and Bottou, Leon},
  booktitle={Proceedings of the 2004 IEEE Computer Society Conference on Computer Vision and Pattern Recognition, 2004. CVPR 2004.},
  volume={2},
  pages={II--104},
  year={2004},
  organization={IEEE}
}

@article{Dicarlo2007,
  title={Untangling invariant object recognition},
  author={DiCarlo, James J and Cox, David D},
  journal={Trends in cognitive sciences},
  volume={11},
  number={8},
  pages={333--341},
  year={2007},
  publisher={Elsevier}
}

@inproceedings{Pinto2011,
  title={Comparing state-of-the-art visual features on invariant object recognition tasks},
  author={Pinto, Nicolas and Barhomi, Youssef and Cox, David D and DiCarlo, James J},
  booktitle={2011 IEEE workshop on Applications of computer vision (WACV)},
  pages={463--470},
  year={2011},
  organization={IEEE}
}

@inproceedings{Torralba2011,
  title={Unbiased look at dataset bias},
  author={Torralba, Antonio and Efros, Alexei A},
  booktitle={CVPR 2011},
  pages={1521--1528},
  year={2011},
  organization={IEEE}
}

@article{Dicarlo2012,
  title={How does the brain solve visual object recognition?},
  author={DiCarlo, James J and Zoccolan, Davide and Rust, Nicole C},
  journal={Neuron},
  volume={73},
  number={3},
  pages={415--434},
  year={2012},
  publisher={Elsevier}
}

@article{Krizhevsky2012,
  title={Imagenet classification with deep convolutional neural networks},
  author={Krizhevsky, Alex and Sutskever, Ilya and Hinton, Geoffrey E},
  journal={Advances in neural information processing systems},
  volume={25},
  year={2012}
}

@article{Simonyan2014,
  title={Very deep convolutional networks for large-scale image recognition},
  author={Simonyan, Karen and Zisserman, Andrew},
  journal={arXiv preprint arXiv:1409.1556},
  year={2014}
}

@article{Chang2015,
  title={Shapenet: An information-rich 3d model repository},
  author={Chang, Angel X and Funkhouser, Thomas and Guibas, Leonidas and Hanrahan, Pat and Huang, Qixing and Li, Zimo and Savarese, Silvio and Savva, Manolis and Song, Shuran and Su, Hao and others},
  journal={arXiv preprint arXiv:1512.03012},
  year={2015}
}

@inproceedings{Szegedy2015,
  title={Going deeper with convolutions},
  author={Szegedy, Christian and Liu, Wei and Jia, Yangqing and Sermanet, Pierre and Reed, Scott and Anguelov, Dragomir and Erhan, Dumitru and Vanhoucke, Vincent and Rabinovich, Andrew},
  booktitle={Proceedings of the IEEE conference on computer vision and pattern recognition},
  pages={1--9},
  year={2015}
}

@article{Alain2016,
  title={Understanding intermediate layers using linear classifier probes},
  author={Alain, Guillaume and Bengio, Yoshua},
  journal={arXiv preprint arXiv:1610.01644},
  year={2016}
}

@inproceedings{Borji2016,
  title={ilab-20m: A large-scale controlled object dataset to investigate deep learning},
  author={Borji, Ali and Izadi, Saeed and Itti, Laurent},
  booktitle={Proceedings of the IEEE Conference on Computer Vision and Pattern Recognition},
  pages={2221--2230},
  year={2016}
}

@inproceedings{He2016,
  title={Deep residual learning for image recognition},
  author={He, Kaiming and Zhang, Xiangyu and Ren, Shaoqing and Sun, Jian},
  booktitle={Proceedings of the IEEE conference on computer vision and pattern recognition},
  pages={770--778},
  year={2016}
}

@inproceedings{Ribeiro2016,
  title={" Why should i trust you?" Explaining the predictions of any classifier},
  author={Ribeiro, Marco Tulio and Singh, Sameer and Guestrin, Carlos},
  booktitle={Proceedings of the 22nd ACM SIGKDD international conference on knowledge discovery and data mining},
  pages={1135--1144},
  year={2016}
}

@inproceedings{Szegedy2016,
  title={Rethinking the inception architecture for computer vision},
  author={Szegedy, Christian and Vanhoucke, Vincent and Ioffe, Sergey and Shlens, Jon and Wojna, Zbigniew},
  booktitle={Proceedings of the IEEE conference on computer vision and pattern recognition},
  pages={2818--2826},
  year={2016}
}

@software{Torchvision2016,
    title        = {TorchVision: PyTorch's Computer Vision library},
    author       = {TorchVision},
    year         = 2016,
    journal      = {GitHub repository},
    publisher    = {GitHub},
    howpublished = {\url{https://github.com/pytorch/vision}}
}

@article{Zhu2016,
  title={Object recognition with and without objects},
  author={Zhu, Zhuotun and Xie, Lingxi and Yuille, Alan L},
  journal={arXiv preprint arXiv:1611.06596},
  year={2016}
}

@inproceedings{Bau2017,
  title={Network dissection: Quantifying interpretability of deep visual representations},
  author={Bau, David and Zhou, Bolei and Khosla, Aditya and Oliva, Aude and Torralba, Antonio},
  booktitle={Proceedings of the IEEE conference on computer vision and pattern recognition},
  pages={6541--6549},
  year={2017}
}

@inproceedings{Huang2017,
  title={Densely connected convolutional networks},
  author={Huang, Gao and Liu, Zhuang and Van Der Maaten, Laurens and Weinberger, Kilian Q},
  booktitle={Proceedings of the IEEE conference on computer vision and pattern recognition},
  pages={4700--4708},
  year={2017}
}

@inproceedings{Johnson2017,
  title={Clevr: A diagnostic dataset for compositional language and elementary visual reasoning},
  author={Johnson, Justin and Hariharan, Bharath and Van Der Maaten, Laurens and Fei-Fei, Li and Lawrence Zitnick, C and Girshick, Ross},
  booktitle={Proceedings of the IEEE conference on computer vision and pattern recognition},
  pages={2901--2910},
  year={2017}
}

@misc{Matthey2017,
author = {Loic Matthey and Irina Higgins and Demis Hassabis and Alexander Lerchner},
title = {dSprites: Disentanglement testing Sprites dataset},
howpublished= {https://github.com/deepmind/dsprites-dataset/},
year = "2017",
}

@article{Baker2018,
  title={Deep convolutional networks do not classify based on global object shape},
  author={Baker, Nicholas and Lu, Hongjing and Erlikhman, Gennady and Kellman, Philip J},
  journal={PLoS computational biology},
  volume={14},
  number={12},
  pages={e1006613},
  year={2018},
  publisher={Public Library of Science San Francisco, CA USA}
}

@inproceedings{Beery2018,
  title={Recognition in terra incognita},
  author={Beery, Sara and Van Horn, Grant and Perona, Pietro},
  booktitle={Proceedings of the European conference on computer vision (ECCV)},
  pages={456--473},
  year={2018}
}

@misc{Burgess2018,
  title={3D Shapes Dataset},
  author={Burgess, Chris and Kim, Hyunjik},
  howpublished={https://github.com/deepmind/3dshapes-dataset/},
  year={2018}
}

@inproceedings{Graziani2018,
  title={Regression concept vectors for bidirectional explanations in histopathology},
  author={Graziani, Mara and Andrearczyk, Vincent and M{\"u}ller, Henning},
  booktitle={International Workshop on Machine Learning in Clinical Neuroimaging},
  pages={124--132},
  year={2018},
  organization={Springer}
}

@inproceedings{Kim2018,
  title={Interpretability beyond feature attribution: Quantitative testing with concept activation vectors (tcav)},
  author={Kim, Been and Wattenberg, Martin and Gilmer, Justin and Cai, Carrie and Wexler, James and Viegas, Fernanda and others},
  booktitle={International conference on machine learning},
  pages={2668--2677},
  year={2018},
  organization={PMLR}
}

@article{Mcinnes2018,
  title={Umap: Uniform manifold approximation and projection for dimension reduction},
  author={McInnes, Leland and Healy, John and Melville, James},
  journal={arXiv preprint arXiv:1802.03426},
  year={2018}
}

@inproceedings{Alcorn2019,
  title = {Strike ({{With}}) a {{Pose}}: {{Neural Networks Are Easily Fooled}} by {{Strange Poses}} of {{Familiar Objects}}},
  shorttitle = {Strike ({{With}}) a {{Pose}}},
  booktitle = {2019 {{IEEE}}/{{CVF Conference}} on {{Computer Vision}} and {{Pattern Recognition}} ({{CVPR}})},
  author = {Alcorn, Michael A. and Li, Qi and Gong, Zhitao and Wang, Chengfei and Mai, Long and Ku, Wei-Shinn and Nguyen, Anh},
  year = 2019,
  pages = {4840--4849},
  publisher = {IEEE},
  address = {Long Beach, CA, USA},
  doi = {10.1109/CVPR.2019.00498},
  copyright = {https://doi.org/10.15223/policy-029},
  isbn = {978-1-7281-3293-8}
}

@article{Barbu2019,
  title={Objectnet: A large-scale bias-controlled dataset for pushing the limits of object recognition models},
  author={Barbu, Andrei and Mayo, David and Alverio, Julian and Luo, William and Wang, Christopher and Gutfreund, Dan and Tenenbaum, Josh and Katz, Boris},
  journal={Advances in neural information processing systems},
  volume={32},
  year={2019}
}

@inproceedings{Geirhos2019,
  title = {{{ImageNet-trained CNNs}} Are Biased towards Texture; Increasing Shape Bias Improves Accuracy and Robustness},
  booktitle = {International {{Conference}} on {{Learning Representations}}},
  author = {Geirhos, Robert and Rubisch, Patricia and Michaelis, Claudio and Bethge, Matthias and Wichmann, Felix A. and Brendel, Wieland},
  year = 2019
}

@inproceedings{Gondal2019,
 author = {Gondal, Muhammad Waleed and Wuthrich, Manuel and Miladinovic, Djordje and Locatello, Francesco and Breidt, Martin and Volchkov, Valentin and Akpo, Joel and Bachem, Olivier and Sch{ö}lkopf, Bernhard and Bauer, Stefan},
 booktitle = {Advances in Neural Information Processing Systems},
 editor = {H. Wallach and H. Larochelle and A. Beygelzimer and F. d\textquotesingle Alch\'{e}-Buc and E. Fox and R. Garnett},
 pages = {},
 publisher = {Curran Associates, Inc.},
 title = {On the Transfer of Inductive Bias from Simulation to the Real World: a New Disentanglement Dataset},
 url = {https://proceedings.neurips.cc/paper/2019/file/d97d404b6119214e4a7018391195240a-Paper.pdf},
 volume = {32},
 year = {2019}
}

@article{Hendrycks2019,
  title={Benchmarking Neural Network Robustness to Common Corruptions and Perturbations},
  author={Dan Hendrycks and Thomas Dietterich},
  journal={Proceedings of the International Conference on Learning Representations},
  year={2019}
}

@inproceedings{Tan2019,
  title={Efficientnet: Rethinking model scaling for convolutional neural networks},
  author={Tan, Mingxing and Le, Quoc},
  booktitle={International conference on machine learning},
  pages={6105--6114},
  year={2019},
  organization={PMLR}
}

@article{Baker2020,
  title={Local features and global shape information in object classification by deep convolutional neural networks},
  author={Baker, Nicholas and Lu, Hongjing and Erlikhman, Gennady and Kellman, Philip J},
  journal={Vision research},
  volume={172},
  pages={46--61},
  year={2020},
  publisher={Elsevier}
}

@article{Dosovitskiy2020,
  title={An image is worth 16x16 words: Transformers for image recognition at scale},
  author={Dosovitskiy, Alexey and Beyer, Lucas and Kolesnikov, Alexander and Weissenborn, Dirk and Zhai, Xiaohua and Unterthiner, Thomas and Dehghani, Mostafa and Minderer, Matthias and Heigold, Georg and Gelly, Sylvain and others},
  journal={arXiv preprint arXiv:2010.11929},
  year={2020}
}

@article{Geirhos2020,
  title = {Shortcut {{Learning}} in {{Deep Neural Networks}}},
  author = {Geirhos, Robert and Jacobsen, J{\"o}rn-Henrik and Michaelis, Claudio and Zemel, Richard and Brendel, Wieland and Bethge, Matthias and Wichmann, Felix A.},
  year = 2020,
  journal = {Nature Machine Intelligence},
  volume = {2},
  pages = {665--673},
  issn = {2522-5839},
  doi = {10.1038/s42256-020-00257-z}
}

@article{Graziani2020,
  title={Concept attribution: Explaining CNN decisions to physicians},
  author={Graziani, Mara and Andrearczyk, Vincent and Marchand-Maillet, St{\'e}phane and M{\"u}ller, Henning},
  journal={Computers in biology and medicine},
  volume={123},
  pages={103865},
  year={2020},
  publisher={Elsevier}
}

@article{Hermann2020,
  title={The origins and prevalence of texture bias in convolutional neural networks},
  author={Hermann, Katherine and Chen, Ting and Kornblith, Simon},
  journal={Advances in neural information processing systems},
  volume={33},
  pages={19000--19015},
  year={2020}
}

@article{Geirhos2021,
  title={Partial success in closing the gap between human and machine vision},
  author={Geirhos, Robert and Narayanappa, Kantharaju and Mitzkus, Benjamin and Thieringer, Tizian and Bethge, Matthias and Wichmann, Felix A and Brendel, Wieland},
  journal={Advances in Neural Information Processing Systems},
  volume={34},
  pages={23885--23899},
  year={2021}
}

@inproceedings{Hendrycks2021,
  title={The many faces of robustness: A critical analysis of out-of-distribution generalization},
  author={Hendrycks, Dan and Basart, Steven and Mu, Norman and Kadavath, Saurav and Wang, Frank and Dorundo, Evan and Desai, Rahul and Zhu, Tyler and Parajuli, Samyak and Guo, Mike and others},
  booktitle={Proceedings of the IEEE/CVF international conference on computer vision},
  pages={8340--8349},
  year={2021}
}

@inproceedings{Liu2021,
  title={Swin transformer: Hierarchical vision transformer using shifted windows},
  author={Liu, Ze and Lin, Yutong and Cao, Yue and Hu, Han and Wei, Yixuan and Zhang, Zheng and Lin, Stephen and Guo, Baining},
  booktitle={Proceedings of the IEEE/CVF international conference on computer vision},
  pages={10012--10022},
  year={2021}
}

@article{Mehrer2021,
  title={An ecologically motivated image dataset for deep learning yields better models of human vision},
  author={Mehrer, Johannes and Spoerer, Courtney J and Jones, Emer C and Kriegeskorte, Nikolaus and Kietzmann, Tim C},
  journal={Proceedings of the National Academy of Sciences},
  volume={118},
  number={8},
  pages={e2011417118},
  year={2021},
  publisher={National Academy of Sciences}
}

@article{Morrison2021,
  title={Exploring corruption robustness: Inductive biases in vision transformers and MLP-mixers},
  author={Morrison, Katelyn and Gilby, Benjamin and Lipchak, Colton and Mattioli, Adam and Kovashka, Adriana},
  journal={arXiv preprint arXiv:2106.13122},
  year={2021}
}

@article{Naseer2021,
  title={Intriguing properties of vision transformers},
  author={Naseer, Muhammad Muzammal and Ranasinghe, Kanchana and Khan, Salman H and Hayat, Munawar and Shahbaz Khan, Fahad and Yang, Ming-Hsuan},
  journal={Advances in Neural Information Processing Systems},
  volume={34},
  pages={23296--23308},
  year={2021}
}

@article{Singla2021,
  title={Salient imagenet: How to discover spurious features in deep learning?},
  author={Singla, Sahil and Feizi, Soheil},
  journal={arXiv preprint arXiv:2110.04301},
  year={2021}
}

@article{Tuli2021,
  title={Are convolutional neural networks or transformers more like human vision?},
  author={Tuli, Shikhar and Dasgupta, Ishita and Grant, Erin and Griffiths, Thomas L},
  journal={arXiv preprint arXiv:2105.07197},
  year={2021}
}

@article{Xiao2021,
  title={Noise or signal: The role of image backgrounds in object recognition},
  author={Xiao, Kai and Engstrom, Logan and Ilyas, Andrew and Madry, Aleksander},
  journal={Proceedings of the International Conference on Learning Representations},
  year={2021}
}

@inproceedings{Zimmermann2021,
  title={Contrastive learning inverts the data generating process},
  author={Zimmermann, Roland S and Sharma, Yash and Schneider, Steffen and Bethge, Matthias and Brendel, Wieland},
  booktitle={International conference on machine learning},
  pages={12979--12990},
  year={2021},
  organization={PMLR}
}

@article{Dong2022,
  title={Viewfool: Evaluating the robustness of visual recognition to adversarial viewpoints},
  author={Dong, Yinpeng and Ruan, Shouwei and Su, Hang and Kang, Caixin and Wei, Xingxing and Zhu, Jun},
  journal={Advances in neural information processing systems},
  volume={35},
  pages={36789--36803},
  year={2022}
}

@software{Jakob2022,
    title = {Mitsuba 3 renderer},
    author = {Wenzel Jakob and Sébastien Speierer and Nicolas Roussel and Merlin Nimier-David and Delio Vicini and Tizian Zeltner and Baptiste Nicolet and Miguel Crespo and Vincent Leroy and Ziyi Zhang},
    note = {https://mitsuba-renderer.org},
    version = {3.8.0},
    year = 2022
}

@inproceedings{Liu2022,
  title={A convnet for the 2020s},
  author={Liu, Zhuang and Mao, Hanzi and Wu, Chao-Yuan and Feichtenhofer, Christoph and Darrell, Trevor and Xie, Saining},
  booktitle={Proceedings of the IEEE/CVF conference on computer vision and pattern recognition},
  pages={11976--11986},
  year={2022}
}

@article{Bordes2023,
  title={Pug: Photorealistic and semantically controllable synthetic data for representation learning},
  author={Bordes, Florian and Shekhar, Shashank and Ibrahim, Mark and Bouchacourt, Diane and Vincent, Pascal and Morcos, Ari},
  journal={Advances in Neural Information Processing Systems},
  volume={36},
  pages={45020--45054},
  year={2023}
}

@article{Deitke2023,
  title={Objaverse-xl: A universe of 10m+ 3d objects},
  author={Deitke, Matt and Liu, Ruoshi and Wallingford, Matthew and Ngo, Huong and Michel, Oscar and Kusupati, Aditya and Fan, Alan and Laforte, Christian and Voleti, Vikram and Gadre, Samir Yitzhak and others},
  journal={Advances in Neural Information Processing Systems},
  volume={36},
  pages={35799--35813},
  year={2023}
}

@article{Garrido2023,
  title={Self-supervised learning of split invariant equivariant representations},
  author={Garrido, Quentin and Najman, Laurent and Lecun, Yann},
  journal={arXiv preprint arXiv:2302.10283},
  year={2023}
}

@inproceedings{Li2023,
  title={Imagenet-e: Benchmarking neural network robustness via attribute editing},
  author={Li, Xiaodan and Chen, Yuefeng and Zhu, Yao and Wang, Shuhui and Zhang, Rong and Xue, Hui},
  booktitle={Proceedings of the IEEE/CVF Conference on Computer Vision and Pattern Recognition},
  pages={20371--20381},
  year={2023}
}

@inproceedings{Venkatesh2023,
  title={Adversarial robustness in discontinuous spaces via alternating sampling \& descent},
  author={Venkatesh, Rahul and Wong, Eric and Kolter, Zico},
  booktitle={Proceedings of the IEEE/CVF Winter Conference on Applications of Computer Vision},
  pages={4662--4671},
  year={2023}
}

@inproceedings{Wu2023,
  title={Omniobject3d: Large-vocabulary 3d object dataset for realistic perception, reconstruction and generation},
  author={Wu, Tong and Zhang, Jiarui and Fu, Xiao and Wang, Yuxin and Ren, Jiawei and Pan, Liang and Wu, Wayne and Yang, Lei and Wang, Jiaqi and Qian, Chen and others},
  booktitle={Proceedings of the IEEE/CVF Conference on Computer Vision and Pattern Recognition},
  pages={803--814},
  year={2023}
}

@article{Lin2024,
  title={The topology and geometry of neural representations},
  author={Lin, Baihan and Kriegeskorte, Nikolaus},
  journal={Proceedings of the National Academy of Sciences},
  volume={121},
  number={42},
  pages={e2317881121},
  year={2024},
  publisher={National Academy of Sciences}
}

@article{Ma2024,
  title={Imagenet3d: Towards general-purpose object-level 3d understanding},
  author={Ma, Wufei and Zhang, Guofeng and Liu, Qihao and Zeng, Guanning and Kortylewski, Adam and Liu, Yaoyao and Yuille, Alan},
  journal={Advances in Neural Information Processing Systems},
  volume={37},
  pages={96127--96149},
  year={2024}
}

@article{Spiess2024,
  title={The sketchfab 3d creative commons collection (s3d3c)},
  author={Spiess, Florian and Waltensp{ü}l, Raphael and Schuldt, Heiko},
  journal={arXiv preprint arXiv:2407.17205},
  year={2024}
}

@inproceedings{Zhang2024,
  title={{ImageNet-D}: Benchmarking neural network robustness on diffusion synthetic object},
  author={Zhang, Chenshuang and Pan, Fei and Kim, Junmo and Kweon, In So and Mao, Chengzhi},
  booktitle={Proceedings of the IEEE/CVF Conference on Computer Vision and Pattern Recognition},
  pages={21752--21762},
  year={2024}
}

@article{Beilharz2025,
  title={MRD: Using Physically Based Differentiable Rendering to Probe Vision Models for 3D Scene Understanding},
  author={Beilharz, Benjamin and Wallis, Thomas SA},
  journal={arXiv preprint arXiv:2512.12307},
  year={2025}
}

@manual{Blender2025,
  title={Blender – a 3D modelling and rendering package},
  author={{Blender Online Community}},
  organization={Blender Foundation},
  year={2025},
  version={4.5.3},
  url={https://www.blender.org},
}

@article{Hesse2025,
  title={Beyond Accuracy: What Matters in Designing Well-Behaved Models?},
  author={Hesse, Robin and Ba{\u{g}}c{\i}, Do{\u{g}}ukan and Schiele, Bernt and Schaub-Meyer, Simone and Roth, Stefan},
  journal={arXiv preprint arXiv:2503.17110},
  year={2025}
}

\clearpage
\appendix
\newpage
\setcounter{section}{0}
\setcounter{figure}{0}
\setcounter{table}{0}

\renewcommand{\thefigure}{A\arabic{figure}}
\renewcommand{\thesection}{A\arabic{section}}
\renewcommand{\thetable}{A\arabic{table}}
\section*{Appendix}

\section{Mesh curation details}
\label{SI:mesh-curation}
The dataset is curated through a fully scripted, modular pipeline that
proceeds from a hand-assembled source list of 3D-asset URLs through to
a set of canonically aligned, scaled, and human-approved meshes
suitable for both rendering and geometric analysis.  The pipeline is
organized into four phases, summarized in
Table~\ref{tab:pipeline-overview} and detailed below.

\begin{table}[h]
\centering
\small
\begin{tabular}{@{}p{0.22\linewidth} p{0.35\linewidth} p{0.37\linewidth}@{}}
\toprule
\textbf{Phase} & \textbf{Goal} & \textbf{Stages} \\
\midrule
1. Provenance and Licensing & Ensure legal compliance and asset authenticity & Validate source list, harvest metadata, filter for redistributable/human-authored assets \\
\addlinespace
2. Acquisition and Integrity & Secure high-quality raw data & Download approved meshes, verify file integrity \\
\addlinespace
3. Mesh-level Curation & Curate asset quality and structural composition & Manual inspection, drop/keep filtering, mesh sub-component splitting \\
\addlinespace
4. Scene Construction & Generate standardized dataset outputs & Alignment, normalization, manual correction, asset and render export \\
\bottomrule
\end{tabular}
\caption{Overview of the four curation phases of the MAPS dataset.}
\label{tab:pipeline-overview}
\end{table}

\subsection{Provenance and licensing}

\paragraph{Source list assembly.}
We started from the complete set of 1000 ImageNet classes
and manually searched Sketchfab for each one in turn, retaining only
\emph{photorealistic} meshes, explicitly excluding stylized,
cartoon, low-poly, abstract, or otherwise non-naturalistic models,
and recording up to eight candidate URLs per class.  Classes for
which no suitable photorealistic asset existed under the relevant
licenses are absent from the final dataset.  The per-class cap of eight is applied at search time rather than at filter time, bounding class size and curation effort while remaining tractable for the subsequent human-review stages.

\paragraph{Metadata harvesting.}
For every validated URL we extract the asset's unique identifier and query the Sketchfab API to obtain its full metadata record (title, author, license, tags, description, geometry statistics). All subsequent licensing, author, AI-generation, and \texttt{noai}-tag filters operate on this metadata record rather than on the manually transcribed fields, which serve only as a search artifact.

\paragraph{Licensing audit.}
An initial pass retained all Creative Commons licenses, including the restricted variants \texttt{CC-BY-NC}, \texttt{CC-BY-ND}, and \texttt{CC-BY-NC-SA}.  We subsequently narrowed the admissible set to \texttt{CC-BY} and \texttt{CC0~Public~Domain} so that the dataset carries a single, uniform redistribution policy: downstream users can incorporate the entire corpus into derivative work without performing a per-mesh license audit. This decision also ensures consistent treatment under non-commercial constraints (e.g.\ paid model-training pipelines), where a single \texttt{CC-BY-NC} mesh would otherwise compromise the whole dataset's usable status. For transparency and to support users with different redistribution requirements, we additionally release the source URLs and metadata for assets under more restrictive Creative Commons licenses (\texttt{CC-BY-NC}, \texttt{CC-BY-ND}, \texttt{CC-BY-NC-SA}) that were excluded from the hosted dataset. These assets are not distributed with MAPS but can be independently accessed and used by researchers whose use case permits it.

\paragraph{Author exclusions.}
Authors whose published terms of use contradict the file-level CC license (e.g.\ a global ``no-redistribution'' clause attached to the account) are excluded by name.

\paragraph{Author-opt-out and AI-generation removal.}
Two further passes remove (i)~assets whose authors carry the explicit \texttt{noai} opt-out tag and (ii)~assets that are themselves AI-generated, identified by tags or descriptions referencing common generative-3D tools. This second pass is essential for ensuring the dataset contains only human-authored geometry and is consistent with the authors' redistribution intent.

\subsection{Acquisition and integrity}

\paragraph{Mesh download.}
Surviving entries are downloaded as glTF-binary (\texttt{.glb}) files into a per-category directory tree. Each asset is identified by a stable shortened hash; collisions across categories are guaranteed unique by widening the prefix length until no two truncations clash.

\paragraph{Download verification.}
A subsequent pass enumerates every retained \texttt{(synset, hash)} pair in the source table and verifies that a corresponding file exists in the download tree, flagging any missing or unexpected
entries.

\subsection{Mesh-level curation}

Raw assets often contain incidental scenery (e.g., display stands or background props) or multiple distinct object instances within a single source file. To ensure dataset purity and granularity, we perform a manual per-mesh review of every asset to:

\begin{enumerate}
\item Isolate and Inspect: We evaluate each component mesh in isolation to verify its quality and relevance to the target category.

\item Categorize Components: We assign each mesh a status of keep, remove, or split. The split operation is used to decouple target objects from environmental geometry or to separate multiple instances of the same category (e.g., several chairs in one scene) into unique dataset entries.

\item Identify Failures: We flag assets that fail to meet quality standards for manual replacement or structural correction.

\end{enumerate}

The decisions JSON is then consumed by a glTF-transform processor that materializes the choices: it removes meshes marked as background, emits one output per ``split'' group while preserving any meshes flagged keep across all groups, and copies through the verbatim asset when no removals were needed.  Single-mesh assets are written as one file per category instance; assets with multiple split groups are written as enumerated variants.  Variant indices therefore enumerate genuinely distinct category instances, not material or color variations of a single object; they are treated as independent entries in all downstream stages. A separate manual-replacement pool is then merged into the processed tree, providing replacements for assets that were unrecoverable through removal alone.

\subsection{Scene construction and standardization}

\paragraph{Cleanup pass.}
A first Blender pass strips imported scaffolding (cameras, lights, empties, armatures, and degenerate placeholder meshes), unparents every remaining mesh while preserving its world transform, and re-parents all meshes under a single root empty at the world origin.

\paragraph{Canonical-frame alignment.}
For each cleaned asset we compute a canonical orientation frame from its pooled vertex cloud.  We use an axis-aligned bounding box (AABB) strategy as follows:
\begin{enumerate}
    \item We sort the world axes by descending AABB extent to assign
          \texttt{(long, sides, up)}.
    \item We resolve the up-axis sign by a gravity prior: the third
          moment of vertex projections along the candidate up-axis
          must be negative (heavier end down).
    \item We enforce a right-handed frame on the sides axis.
    \item For each category, we set the first processed asset as the
          class anchor.  Subsequent assets within the same
          category compare a one-dimensional mass-profile histogram
          along the long axis to the anchor and apply a $180^{\circ}$
          flip when the mirrored profile fits better. This produces
          consistent front/back orientation across instances of a
          class.
\end{enumerate}

\paragraph{Centering and scale normalization.}
Each asset is centered at the world origin using its area-weighted surface centroid,
\begin{equation*}
    \mathbf{c} \;=\; \frac{\sum_{f \in F} A_f \, \mathbf{c}_f}{\sum_{f \in F} A_f},
\end{equation*}
where $A_f$ and $\mathbf{c}_f$ denote the area and centroid of face $f$.  This is unbiased by tessellation density, robust to lopsided geometry, and well-defined for non-watertight meshes.  After centering, every asset is uniformly rescaled so that its furthest vertex sits on a sphere of radius $r = 0.5$,
\begin{equation*}
    s \;=\; \frac{r}{\max_{v \in V} \lVert v - \mathbf{c} \rVert},
\end{equation*}
yielding vertex coordinates that are directly comparable across the
dataset.

\paragraph{Manual orientation review.}
The intra-class anchor procedure aligns most assets correctly, but a residual fraction requires explicit Euler correction (typically $90^{\circ}$ or $180^{\circ}$ rotations about the up axis). We built a lightweight web tool that lists every aligned asset, converts each scene to glTF on demand, and lets a human curator dial an Euler-XYZ correction. Approved rotations are baked into vertex coordinates (preserving UV maps and material bindings).  %

\paragraph{Asset export and manifest synchronization.}
The approved set is finally projected back to the source glTF format by copying the original meshes (filtered by retention).  Two global manifests, a TSV inventory and a JSON metadata record, are synchronized to the retained UIDs so downstream code can operate on the curated subset alone.

\paragraph{Visual quality assurance.}
For visual review we render every retained scene to a fixed resolution PNG using a physically based renderer.  An in-memory rescale option allows the unit sphere to be visualized at any chosen apparent radius without altering on-disk geometry, enabling consistent thumbnail grids for rapid manual inspection.

\section{Parameter space of scene factors}
\label{SI:parameter_space}

\begin{figure}[t!]
  \centering
\begin{adjustbox}{max width=\textwidth}
  \includegraphics{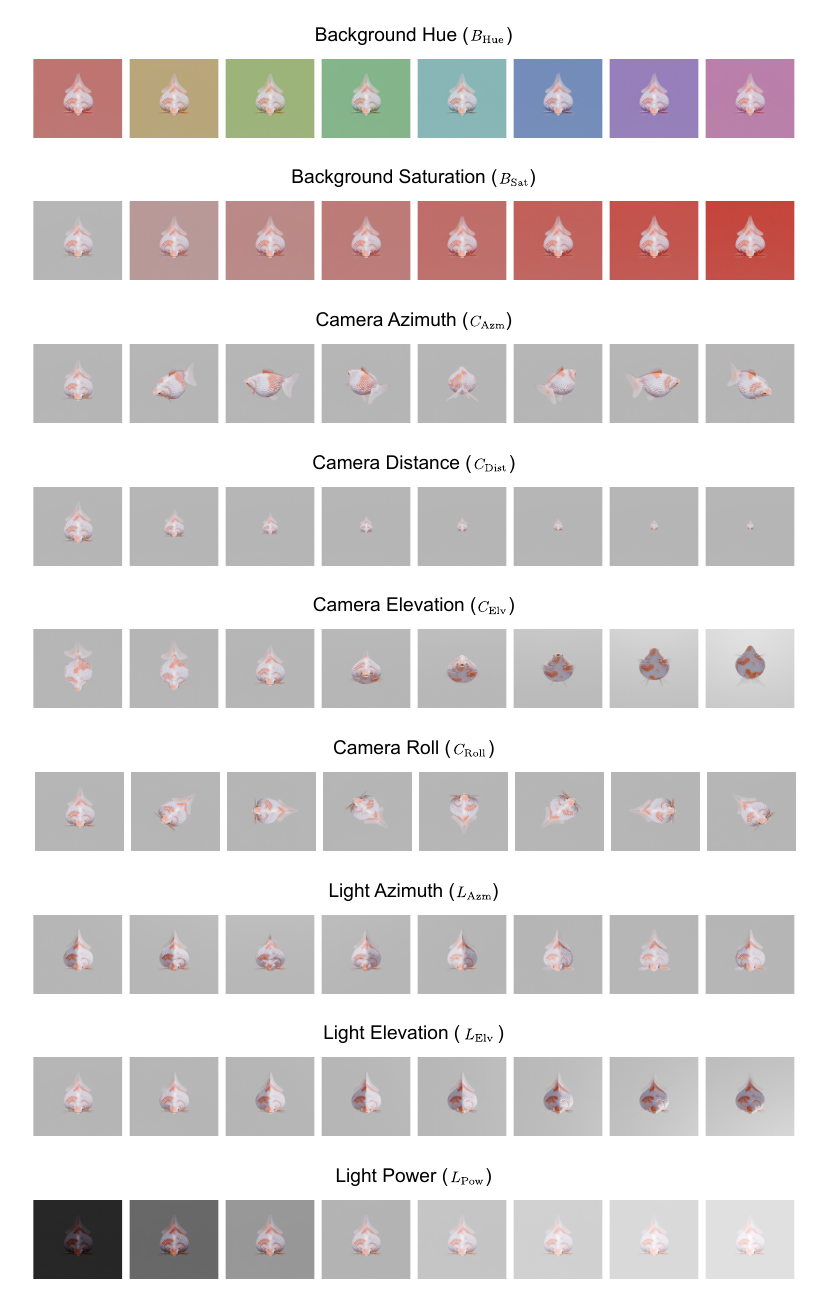}
  \end{adjustbox}
  \caption{\textbf{Visualization of the MAPS scene-parameter space.} Each row shows renderings of the same sample mesh (goldfish class) as a single scene parameter is varied monotonically across its full range (see Table~\ref{tabSI:search_space}) while the remaining eight parameters are held fixed.} %
  \label{figSI:goldfish-param-space}
\end{figure}

Each scene is parameterized by a $9$-dimensional vector
$\boldsymbol{\theta}$ comprising camera pose (azimuth, elevation, distance,
roll), illumination (azimuth, elevation, power), and background appearance (hue, saturation). The bounds on each dimension are listed in Table~\ref{tabSI:search_space}. The remaining scene properties, namely object identity, mesh-native surface materials and textures, single-object placement at the scene origin, and a single directional light source, are held fixed. Fig.~\ref{figSI:goldfish-param-space} shows example renderings under variation of each parameter while the others are held fixed. %
For the validation in Section~\ref{sec:dataset} all dimensions are linearly normalized to the unit interval prior to optimization. Angular parameters are additionally encoded by their sine and cosine components for regression analyses to preserve cyclic structure.

\begin{table}[H]
    \caption{\textbf{Scene parameters.} The MAPS dataset includes 9 different scene parameters that can be systematically manipulated to generate a wide variety of visual stimuli. Each parameter is associated with a specific range of values, allowing for controlled exploration of the visual space.}
    \centering
    \vspace{0.3cm} %
        \begin{tabular}{cccc}
            \toprule
                \textbf{Symbol} & \textbf{Scene Element} & \textbf{Property}  & \textbf{Range} \\
            \midrule
                $B_{\text{Hue}}$       & Background        & Hue                     & $[0, 2\pi)$    \\ 
                $B_{\text{Sat}}$       & Background        & Saturation              & $[0, 1]$       \\
                $C_{\text{Azm}}$       & Camera            & Azimuth                 & $[0, 2\pi)$    \\
                $C_{\text{Dist}}$       & Camera            & Distance                & $[1, 8]$       \\
                $C_{\text{Elv}}$       & Camera            & Elevation               & $[0, \pi]$     \\
                $C_{\text{Roll}}$       & Camera            & Roll                    & $[0, 2\pi)$    \\
                $L_{\text{Azm}}$       & Light             & Azimuth                 & $[0, 2\pi)$    \\
                $L_{\text{Elv}}$       & Light             & Elevation   	         & $[0, \pi]$     \\
                $L_{\text{Pow}}$      & Light              & Power               & $[0.1, 1]$     \\
            \bottomrule
        \end{tabular}
        \label{tabSI:search_space}
\end{table}

\section{Scene Optimization Pipeline}
\label{SI:cmaes}

We use Covariance Matrix Adaptation Evolution Strategy (CMA-ES)~\cite{Hansen2001} to identify scene rendering parameters under which a pretrained ImageNet classifier correctly recognizes each $3$D object. Related work has explored CMA-ES for adversarial search in discontinuous scene-parameter spaces \citep{Venkatesh2023}; here we use it to find any configuration yielding correct classification rather than a worst-case failure. The procedure is applied independently to every scene to confirm that each mesh is recognizable under at least one scene configuration. The search space is defined as in Table~\ref{tabSI:search_space}.

\paragraph{Objective Function.}
For each mesh, CMA-ES searches the scene-parameter space $\boldsymbol{\theta} \in \Theta$ (the nine-dimensional space defined as in Table~\ref{tabSI:search_space}) to find a configuration whose rendered image $I(\boldsymbol{\theta})$ is classified as the target class $c$ by a reference model $\mathcal{M}$. The objective function is defined as the model's log-probability of the target class:

\begin{equation*}
\mathcal{L}(\boldsymbol{\theta} \mid c, \mathcal{M}) = z_c(I(\boldsymbol{\theta}); \mathcal{M}) - \log \sum_{k=1}^{K} \exp\bigl(z_k(I(\boldsymbol{\theta}); \mathcal{M})\bigr)
\label{eq:objective}
\end{equation*}

where $z_k(I(\boldsymbol{\theta}); \mathcal{M})$ is the $k$-th class logit produced for the rendered image, and $K=1000$ for the ImageNet-1k dataset. By construction, $\mathcal{L}$ equals the log of the target-class softmax probability, so maximizing $\mathcal{L}$ drives the model's posterior mass towards class $c$. Since CMA-ES is a minimization algorithm, we define the fitness function $\phi$ as the negative log-probability:

\begin{equation*}
\phi(\boldsymbol{\theta} \mid c, \mathcal{M}) = -\mathcal{L}(\boldsymbol{\theta} \mid c, \mathcal{M}),
\label{eq:fitness}
\end{equation*}

so that minimizing $\phi$ implies maximizing the target-class log-probability.

\paragraph{Optimization procedure.}
The optimizer is initialized from a uniformly sampled point in $\Theta$ with initial step
size $\sigma_0 = 0.2$. The population size follows the default CMA-ES
heuristic ($\lambda = 4 + \lfloor 3 \ln d \rfloor$, giving $\lambda = 6$ for
$d = 9$). At each generation, candidate parameter vectors are decoded into
scene configurations, rendered with Blender's Cycles engine at
$224 \times 224$ resolution, and forwarded through the classifier in a single
batched pass.

\paragraph{Resilience to initialization.}
A single CMA-ES run may converge to a local optimum that fails to classify the object correctly. To mitigate this, each scene is allocated up to $K = 10$ independent restarts with distinct random seeds drawn deterministically from a global initial seed. Restarts proceed sequentially: as soon as one restart satisfies the stopping criterion, the remaining budget is forfeited and we move to the next scene. A scene is declared a \emph{failure} only if all $K$ restarts exhaust the iteration budget without producing a correct classification.

\paragraph{Failure documentation.}
For the full MAPS dataset, 5 out of 2,618 meshes fail the CMA-ES recognizability criterion across all three reference models (AlexNet, ResNet-50, and ViT-B/16), meaning that none of the $K=10$ restarts finds a scene configuration that yields correct top-1 classification within the 100-generation budget. We repeated the analysis with the camera constrained to a distance between 0 and 1 from the object, yielding correct classification for the five meshes. The original distance from 1 to 8 was constrained to have a minimum value 1 to avoid crossings between the camera and the meshes. However, for elongated objects this prevents the camera from being close enough for the mesh to be recognized, also revealing a tendency of models to rely on short distances to make correct predictions (Fig. ~\ref{figSI:cmaes-failure-models}).

\begin{table}[H]
\centering
\caption{\textbf{Hyperparameters of the CMA-ES-based optimization pipeline.}}
\vspace{0.3cm}
\begin{tabular}{lll}
\toprule
\textbf{Symbol} & \textbf{Description} & \textbf{Value} \\
\midrule
$T_{\max}$ & Maximum CMA-ES generations per restart & $100$ \\
$\sigma_0$ & Initial CMA-ES step size                & $0.2$ \\
$\lambda$  & Population size                         & CMA-ES default ($\lambda = 6$) \\
$K$        & Maximum restarts per scene              & $10$ \\
$d$        & Search-space dimensionality             & $9$ \\
$c$        & Target ImageNet class                   & per-synset, from labels \\
\bottomrule
\end{tabular}
\label{tab:hyperparameters}
\end{table}

\section{Vision models probed with MAPS}
\label{SI:vision_models}

We probe a total of 20 vision models varying over architecture family (CNN, Transformer) and parameter count. The full list with reported ImageNet-1k validation accuracies is in Table~\ref{tab:breakdown-probed-models}. For all models we use the V1 checkpoints from TorchVision \footnote{\url{https://github.com/pytorch/vision}}, pretrained on ImageNet-1k.

\begin{table}[H]
    \caption{\textbf{Pretrained vision models probed with MAPS.} Architecture family, training objective, training dataset, and reported top-1 accuracy on the ImageNet-1k validation set, for the 20 models evaluated in this study. All checkpoints are publicly available via TorchVision V1.}%
    \label{tab:breakdown-probed-models}
    \centering
    \footnotesize
    \vspace{0.3cm} %
        \begin{tabular}{lccccc}
            \toprule
                \textbf{Model name} & \textbf{Family} & \makecell{\textbf{Training} \\ \textbf{objective}} & \makecell{\textbf{Training} \\ \textbf{data}} & \textbf{\# Param.} & \makecell{\textbf{INet-1k} \\ \textbf{Acc@1}} \\
            \midrule
                AlexNet \citep{Krizhevsky2012} & CNN         & supervised        &   INet-1k & 61.1M  & 56.52 \\ 
                ConvNeXt-Base \citep{Liu2022} & CNN         & supervised  & INet-1k & 88.6M & 84.06\\
                ConvNeXt-Small \citep{Liu2022} & CNN         & supervised  & INet-1k & 50.2M & 83.61 \\
                ConvNeXt-Tiny \citep{Liu2022} & CNN         & supervised  & INet-1k & 28.6M & 82.52 \\
                DenseNet-121 \citep{Huang2017} & CNN         & supervised   & INet-1k & 8.0M & 74.43\\ 
                DenseNet-161 \citep{Huang2017} & CNN         & supervised   & INet-1k  & 28.7M & 77.14 \\
                DenseNet-201 \citep{Huang2017} & CNN         & supervised   & INet-1k  & 20M  & 76.90  \\
                EfficientNet-B0 \citep{Tan2019}  & CNN         & supervised   & INet-1k  & 5.3M & 77.69 \\
                GoogLeNet \citep{Szegedy2015} & CNN         & supervised  & INet-1k & 6.6M & 69.78\\ 
                Inception-v3 \citep{Szegedy2016} & CNN         & supervised & INet-1k & 27.2M & 77.29 \\ 
                ResNet-18 \citep{He2016} & CNN         & supervised   & INet-1k  & 11.7M & 69.76 \\
                ResNet-50 \citep{He2016} & CNN         & supervised & INet-1k  & 25.6M & 76.13\\ 
                ResNet-101 \citep{He2016} & CNN         & supervised   & INet-1k  & 44.5M & 77.37  \\ 
                VGG-16 \citep{Simonyan2014} & CNN         & supervised  & INet-1k & 138.4M & 71.59\\
                VGG-19 \citep{Simonyan2014} & CNN         & supervised  & INet-1k & 143.7M & 72.38 \\
                ViT-B/16 \citep{Dosovitskiy2020} & Transformer         & supervised  & INet-1k & 86.6M & 81.07\\
                ViT-L/16 \citep{Dosovitskiy2020} & Transformer         & supervised  & INet-1k & 304.3M & 79.66\\
                Swin-B \citep{Liu2021} & Transformer         & supervised  & INet-1k & 87.8M & 83.58\\
                Swin-S \citep{Liu2021} & Transformer         & supervised  & INet-1k & 49.6M & 83.20\\ 
                Swin-T \citep{Liu2021} & Transformer         & supervised  & INet-1k & 28.3M & 81.47\\
                
            \bottomrule
        \end{tabular}
\end{table}

\section{Regression-Based Analysis}
\label{SI:regression}

\paragraph{Linear Regression.}
We approximate the model performance as a linear
function of the scene parameters, allowing us to estimate the
sensitivity of models for a given object. Let $\mathbf{X}$ be the
matrix of scene parameters across images and $\mathbf{y}$ the vector
of model performance per image, the model is
\begin{equation*}
    \mathbf{y} \;=\; \mathbf{X}\boldsymbol{\beta} + \boldsymbol{\epsilon},
\end{equation*}
and we estimate $\boldsymbol{\beta}$ by ordinary least squares,
\begin{equation*}
    \boldsymbol{\beta}^{*}
    \;=\; \arg\min_{\boldsymbol{\beta}}\,
    \| \mathbf{X}\boldsymbol{\beta} - \mathbf{y} \|^{2}
    \;=\; (\mathbf{X}^{\top} \mathbf{X})^{-1}\,\mathbf{X}^{\top} \mathbf{y}.
\end{equation*}
When both $\mathbf{X}$ and $\mathbf{y}$ are standardized to zero mean
and unit variance, $\mathbf{X}^{\top} \mathbf{X}$ is proportional to the
correlation matrix of the scene parameters, $\mathbf{R}$, and
$\mathbf{X}^{\top} \mathbf{y}$ to the vector of correlations between the
parameters and the response, $\mathbf{r}$, so that
$\boldsymbol{\beta}^{*} = \mathbf{R}^{-1} \mathbf{r}$. When the
parameters are uncorrelated ($\mathbf{R} \approx \mathbf{I}$), as is
approximately the case here since they are sampled from a Latin
hypercube~\citep{Mckay1979}, the weights reduce to the direct
correlations, $\boldsymbol{\beta}^{*} \approx \mathbf{r}$, providing an
intuitive measure of sensitivity.

The fit is evaluated by $10$-fold cross-validation: in each fold the
columns of $\mathbf{X}$ and the target $\mathbf{y}$ are standardized on
the training split alone, $\boldsymbol{\beta}^{*}$ is refit, and we
report the coefficient of determination on the held-out split,
\begin{equation*}
    R^{2}
    \;=\; 1 \;-\;
    \frac{\sum_{i\in\text{test}}(y_i - \hat{y}_i)^{2}}
         {\sum_{i\in\text{test}}(y_i - \bar{y})^{2}},
\end{equation*}
averaged across folds. For circular parameters (background hue, camera
azimuth, camera roll, light azimuth) we encode each angle $\theta$ as a
$(\cos\theta,\sin\theta)$ pair, and report a single weight per scene
parameter as $|\boldsymbol{\beta}_{p}^{*}| = \sqrt{\beta_{p,\cos}^{2} +
\beta_{p,\sin}^{2}}$.

\paragraph{Nonlinear Analysis.}
To capture nonlinear effects, we extend the regressor with degree-two
polynomial features, including squared terms and pairwise interactions
between scene parameters. Circular parameters are encoded as
$(\cos\theta,\sin\theta)$ pairs before the polynomial expansion,
so that their squares and $\cos\cdot\sin$ cross-products are also
included. Stacking the linear, quadratic and interaction terms into a
single design matrix $\mathbf{Z}$, the model becomes
\begin{equation*}
    \mathbf{y} \;=\; \mathbf{Z}\boldsymbol{\gamma} + \boldsymbol{\epsilon}.
\end{equation*}
Because $\mathbf{Z}$ contains many strongly collinear columns, ordinary
least squares is unstable; we therefore estimate $\boldsymbol{\gamma}$
by ridge regression,
\begin{equation*}
    \boldsymbol{\gamma}^{*}
    \;=\; \arg\min_{\boldsymbol{\gamma}}\,
    \| \mathbf{Z}\boldsymbol{\gamma} - \mathbf{y} \|^{2}
    \;+\; \alpha\, \|\boldsymbol{\gamma}\|^{2}
    \;=\; (\mathbf{Z}^{\top}\mathbf{Z} + \alpha\,\mathbf{I})^{-1}\,
          \mathbf{Z}^{\top}\mathbf{y},
\end{equation*}
with $\alpha$ selected per training fold by generalized cross-validation
over the logarithmic grid $\{10^{-2},10^{-1},\dots,10^{4}\}$.

The fit is evaluated by $10$-fold cross-validation using the same
protocol as in the linear case: in each fold the columns of $\mathbf{Z}$
and the target $\mathbf{y}$ are standardized on the training split
alone, $\alpha$ and $\boldsymbol{\gamma}^{*}$ are refit, and we report
the held-out $R^{2}$ averaged across folds. Because $\mathbf{Z}$ now
contains correlated columns, individual entries of
$\boldsymbol{\gamma}^{*}$ are no longer interpretable as marginal
correlations. To obtain a sensitivity score per scene parameter we
therefore aggregate coefficients across every polynomial feature that
involves it: for each parameter $p$,
\begin{equation*}
    w_p \;=\; \sum_{f\in\mathcal{F}(p)} |\gamma^{*}_{f}|,
\end{equation*}
where $\mathcal{F}(p)$ is the set of polynomial features (linear,
square or product) that depend on $p$. We further split $w_{p}$ into a
main-effect term, features whose only input is $p$ including its
square, and an interaction term collecting cross-products with another
parameter; for circular variables the $\cos$ and $\sin$ design columns
are folded together so that exactly one weight is reported per scene
parameter.

\section{Compute resources}
\label{SI:compute_resources}

Computational tasks were distributed based on hardware requirements:

\begin{itemize}
    \item  Optimization and Rendering: To accelerate the rendering and optimization pipelines, we utilized four NVIDIA A100-SXM4-80GB GPUs and four NVIDIA RTX A6000 GPUs. 

\item Statistical Analysis: All regression analyses and data post-processing, including the figures, were performed on a MacBook Pro equipped with an Apple M4 chip and 32GB of unified memory.
\end{itemize}

\section{Dataset Availability}
\label{SI:dataset_availability}

MAPS and all associated code, including the rendering pipeline, curation pipeline, and figure reproduction scripts, will be publicly available upon publication of this manuscript.

\begin{landscape}
    \thispagestyle{lscape_footer} %
    
    \vspace*{\fill} %
    \begin{center}
        \captionof{table}{\textbf{Full Cluster Analysis Overview.}}
        \label{tab:full_cluster}
        \vspace{0.5cm}
        
        \begin{adjustbox}{max width=0.9\linewidth, max totalheight=0.7\textheight, center}
            \begin{tabular}{ccccc}
                \toprule
                \textbf{Num.} & \textbf{Cluster Name} & \textbf{WordNet Root} & \textbf{Rep. Class} & \textbf{Neighbors} \\
                \midrule
                1                     & Large Structures and Landmarks              & structure.n.01                & Spiral (506)                   & Fountain (562), Obelisk (682), Lighthouse (437), Triumphal Arch (873)   \\ 
                2                     & Protective Gear \& Personal Accessories     & covering.n.02                 & Ring-Binder (446)              & Bell Tower (442), Tile Roof (858), Birdhouse (448), Thimble (855)       \\
                3                     & Home \& Kitchen Appliances                  & home\_appliance.n.01          & Clothes Iron (606)             & Sewing Machine (606), Vacuum Cleaner (882), Dishwasher (534), Microwave (651) \\
                4                     & Clothing \& Headwear                        & clothing.n.01          & Suit (834)                     & Abaya (399), Scarf (824), T-Shirt (610), Jeans (608) \\
                5                     & Musical Instruments                         & musical\_instrument.n.01      & Pipe Organ (687)               & Accordion (401), Ocarina (684), Trombone (875), Pan Flute (699)  \\
                6                     & Technical Instruments \& Weapons            & instrument.n.01               & Barometer (426)                & Weighing Scale (778), Ruler (769), Sundial (835), Analog Clock (409)  \\
                7                     & Mechanical \& Electronic Hardware           & device.n.01                   & Lighter (626)                  & Remote Control (761), Plectrum (714). Oil-Filter (686), Typewriter Keyboard (878)  \\
                8                     & Maritime \& Aerial Transport                & craft.n.02                    & Gondola (576)                  & Fireboat (554), Sailboat (914), Motorboat (814), Canoe (472)  \\
                9                     & Ground Vehicles \& Automobiles              & self-propelled\_vehicle.n.01  & Golf Cart (575)                & Snowplow (803), Go-Kart (573), Ambulance (407), Tow Truck (864)  \\
               10                     & Containers \& Storage Vessels               & instrumentality.n.03          & Cassette (481)                 & Plant Pot (738), Envelope (549), Mailbox (637), Measuring Cup (647)  \\
               11                     & Furniture                                   & furniture.n.01                & Wardrobe (894)                 & Bookcase (453), Commode (493), China Cabinet (495), Cradle (516)  \\
               12                     & Personal Tech \& Fitness Gear               & equipment.n.01                & Baseball (429)                 & Volleyball (890), Soccer Ball (805), Basketball (430), Punching Bag (747)  \\
               13                     & Manual Tools \& Elongated Objects           & artifact.n.01                 & Drumstick (542)                & Matchstick (644), Crutch (523), Flagpole (557), Racket (752)  \\
               14                     & Miscellaneous            & entity.n.01                   & Pizza (963)                    & Burrito (965), Hotdog (934), Cheeseburger (933), Plate (923)  \\
               15                     & Fruits \& Seeds                             & fruit.n.01                    & Acorn (988)                    & Fig (952), Pineapple (953), Banana (954), Custard Apple (956)  \\
               16                     & Vegetables \& Fungi                         & vegetable.n.01                & Cucumber (943)                 & Artichoke (944), Mushroom (947), Broccoli (937), Cauliflower (938)  \\
               17                     & Invertebrates \& Small Creatures            & organism.n.01                 & Trilobite (69)                 & Centipede (79), Scorpion (71), Long-Horned Beetle (303), Tiger Beetle (300)  \\
               18                     & Mammals                                     & mammal.n.01                   & Hyena (276)                    & Chihuahua (151), Pomeranian (259), Pembroke (263), Polar bear (296)  \\ 
               19                     & Fish \& Aquatic Animals                     & fish.n.01                     & Eel (390)                      & Goldfish (1), Sturgeon (394), Pufferfish (397), Clownfish (393) \\
               20                     & Birds, Reptiles \& Amphibians               & vertebrate.n.01               & Hen (8)                        & Ostrich (9), Macaw (88), Bee Eater (92), Hummingbird (94) \\
                \bottomrule
            \end{tabular}
        \end{adjustbox}
    \end{center}
    \vspace*{\fill} %
\end{landscape}

\begin{figure}
  \centering
\begin{adjustbox}{max width=\textwidth}
  \includegraphics{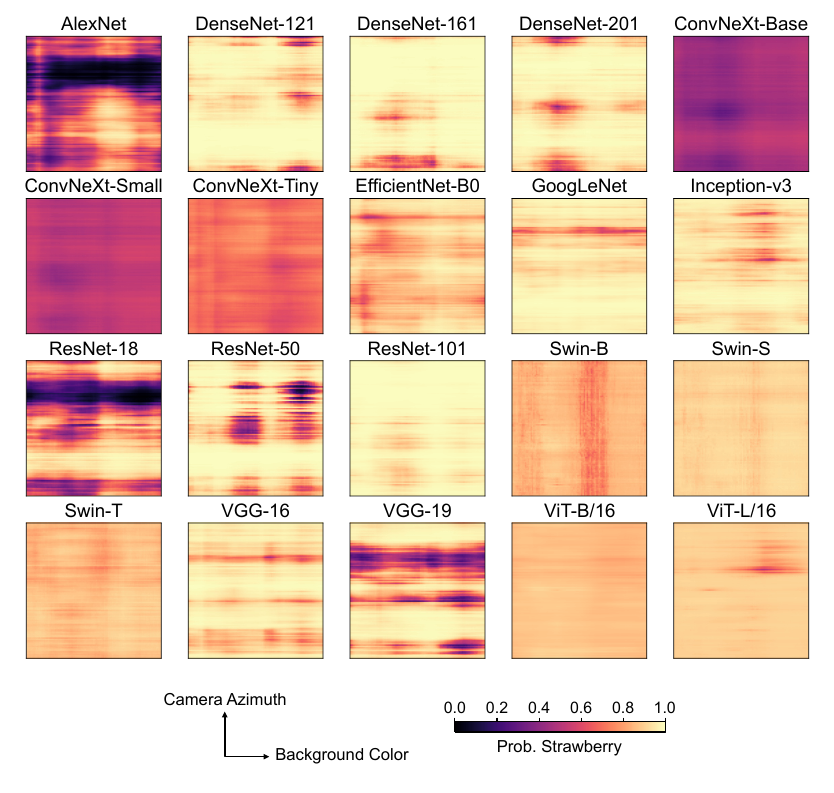}
  \end{adjustbox}
  \caption{\textbf{Confidence regions for joint sweeping of background hue and camera azimuth of a strawberry mesh.} For a single strawberry mesh, we rendered 10,000 images under joint variation of two circular scene factors (background hue and camera azimuth, 100 steps each across their whole range) with all other scene factors held fixed at their default parameters. Each rendered image was passed through a vision model, and the confidence for the correct class was extracted (color-coded).}
  \label{figSI:confidence_models_2D}
\end{figure}

\newpage
\begin{landscape}
    \thispagestyle{lscape_footer} %

\begin{figure}
  \centering
\begin{adjustbox}{max width=1.7\textwidth}
  \includegraphics{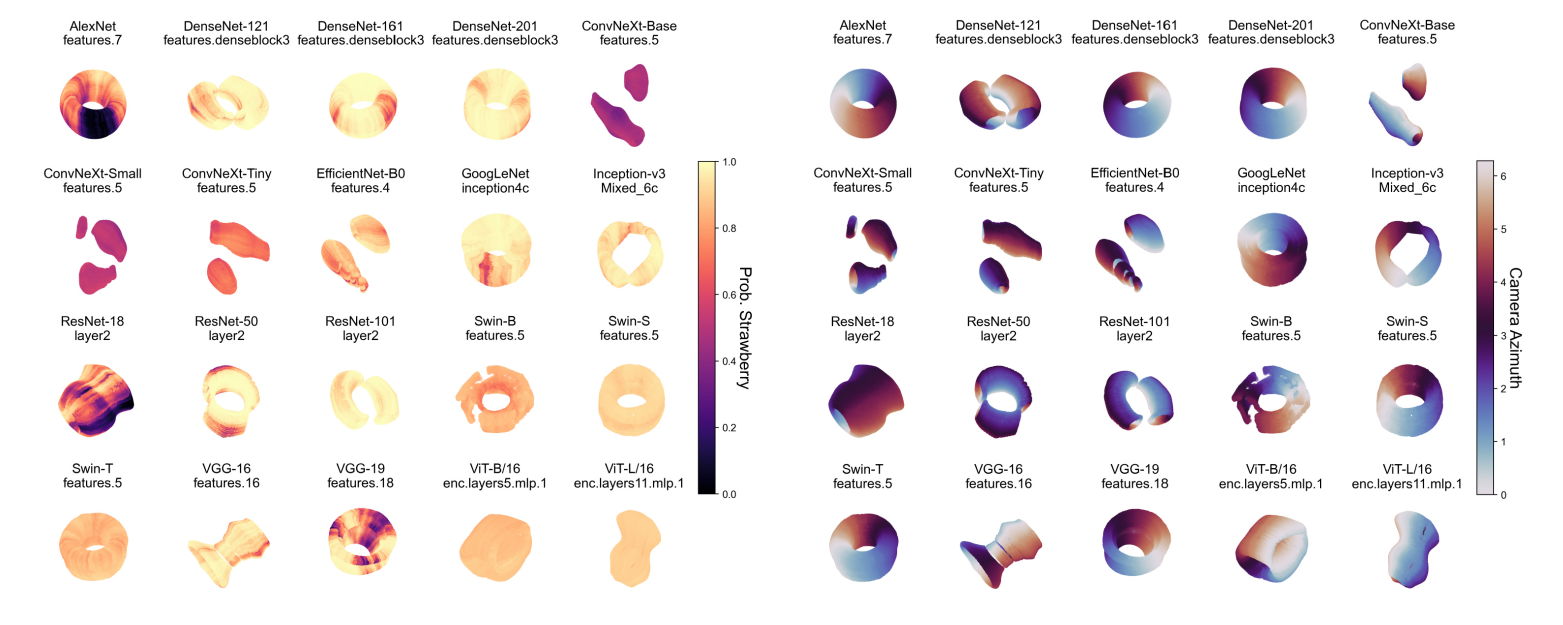}
  \end{adjustbox}
  \caption{\textbf{Activation manifolds for the vision models for joint sweeping of background hue and camera azimuth.} For a single strawberry mesh, we rendered 10,000 images under joint continuous variation of two circular scene parameters (background hue and camera azimuth) with all other scene factors held fixed. The product of these two circular factors defined a toroidal parameter manifold. We subsampled 5,000 images and passed each through the different pretrained models then intermediate activations were extracted (specific layer specified above each panel). For each model, activations were reduced to the first 10 principal components with highest explained variance and finally projected to three dimensions with UMAP \cite{Mcinnes2018}. Each point corresponds to one rendered image and is color-coded by the model's softmax probability assigned to the correct class (left) and the parameter of the camera azimuth (right).}
  \label{figSI:activation-manifolds}
\end{figure}

\end{landscape}

\begin{figure}
  \centering
\begin{adjustbox}{max width=\textwidth}
  \includegraphics{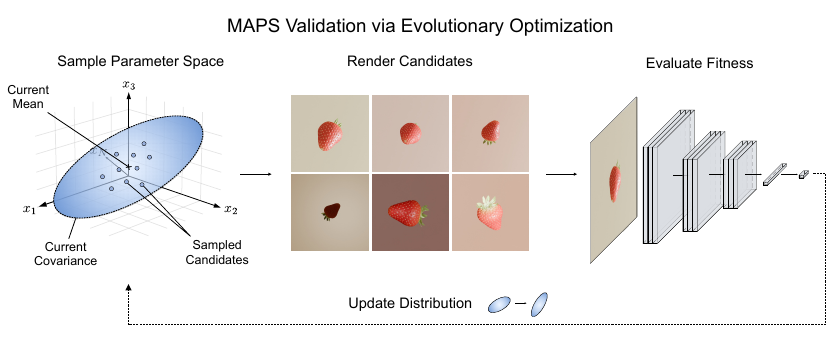}
  \end{adjustbox}
  \caption{\textbf{Validating MAPS objects under optimized scene parameters.} For each mesh, CMA-ES samples candidate scene parameters, renders the corresponding images, evaluates them with one of the models, and updates the sampling distribution toward configurations with higher target-class probability until the object is correctly classified or the maximum number of generations (100) is reached.}
  \label{figSI:optimization-pipeline}
\end{figure}

\begin{figure}
 \centering
\begin{adjustbox}{max width=\textwidth}
 \includegraphics{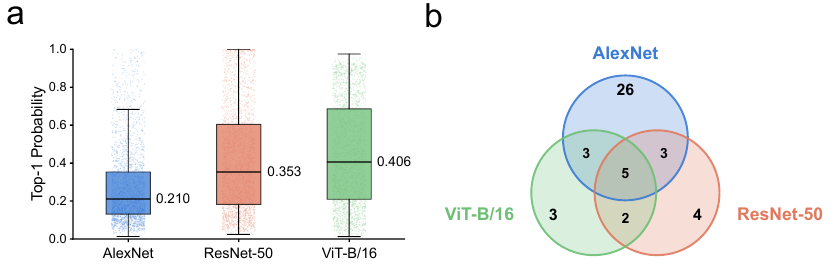}
 \end{adjustbox}
 \caption{\textbf{CMA-ES recognizability validation across architectures.} We performed the per-mesh CMA-ES validation procedure using three reference classifiers from distinct architecture families: AlexNet, ResNet-50 and ViT-B/16. \textbf{(a) Top-1 probability at the best CMA-ES configuration per mesh, for correctly recognized meshes.} \textbf{(b) Overlap of unrecognized meshes across models.} Five meshes fail all three models under the full parameter range (center of Venn diagram). AlexNet accounts for a disproportionate share of unique failures (26 meshes unrecognized) compared to ResNet-50 and ViT-B/16, consistent with its lower median recognition confidence.}
 \label{figSI:cmaes-statistics}
\end{figure}

\begin{figure}
  \centering
\begin{adjustbox}{max width=\textwidth}
  \includegraphics{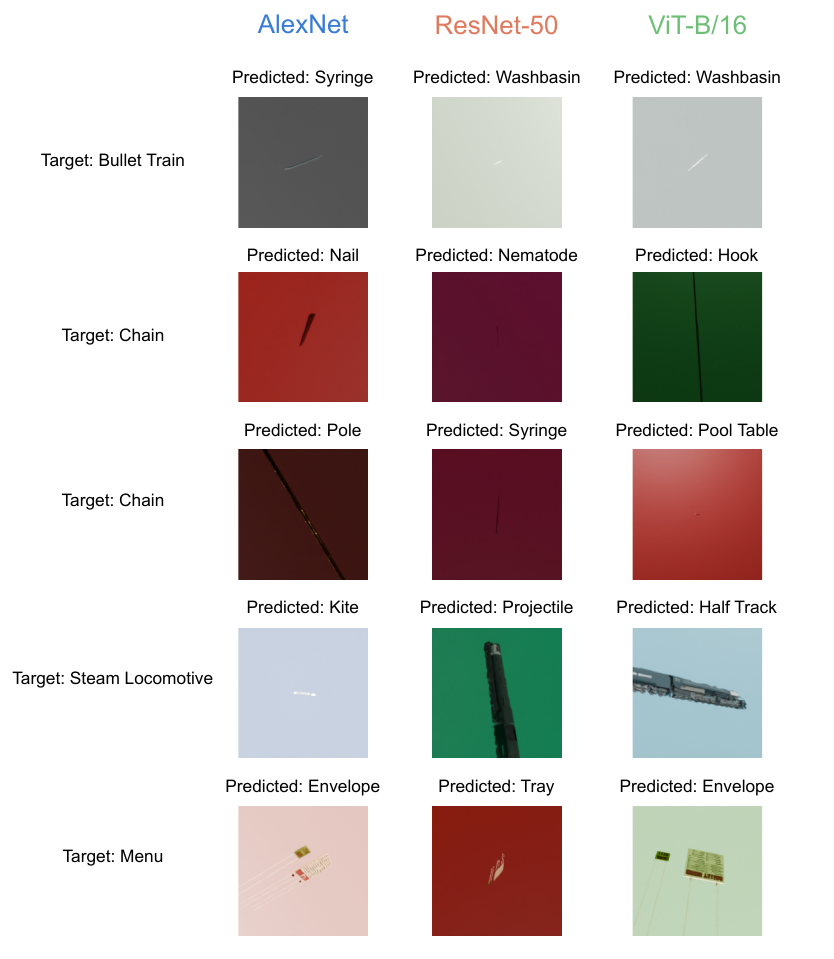}
  \end{adjustbox}
  \caption{\textbf{All CMA-ES failure cases for all three models AlexNet, ResNet-50 and ViT-B/16 across the full MAPS dataset (5 out of 2,618 meshes).} Each row shows one target mesh for which all $K=10$ restarts exhausted their budget without producing a correct top-1 classification. For each model, we show the best solution after the ten restarts.}
  \label{figSI:cmaes-failure-models}
\end{figure}

\begin{figure}
  \centering
\begin{adjustbox}{max width=\textwidth}
  \includegraphics{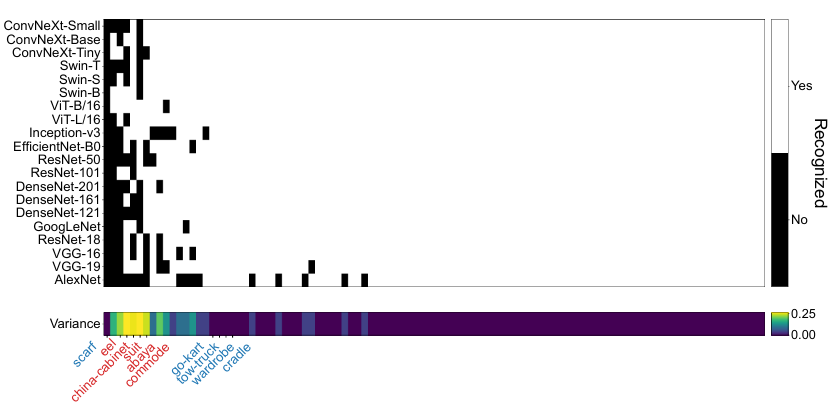}
  \end{adjustbox}
    \caption{\textbf{Maximum top-1 accuracy heatmap across models (rows) and the 100 evaluated classes from MAPS (columns).} Models are sorted by mean accuracy across classes (bottom to top), and classes are sorted by mean accuracy across models (left to right). The bottom strip shows the per-class recognition variance across models. Labels are displayed for the classes with the top-5 highest (red) and lowest (blue) recognition variance across models. Scarf is the only class not recognized by any model.}
    \label{figSI:max-acc-lhs-maps}
\end{figure}

\begin{figure}
  \centering
\begin{adjustbox}{max width=\textwidth}
  \includegraphics{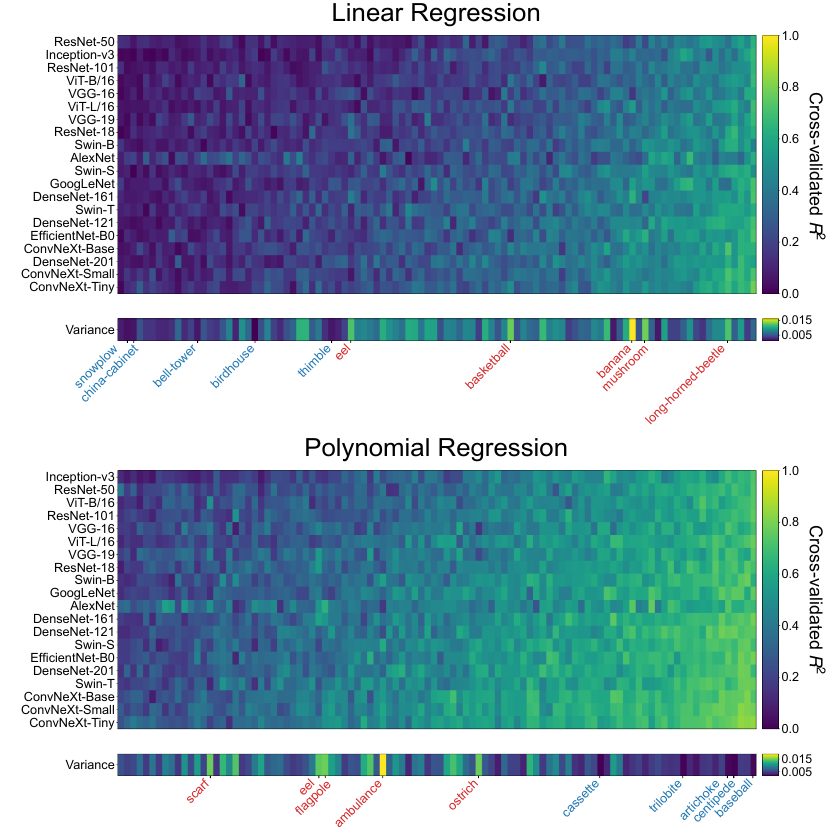}
  \end{adjustbox}
  \caption{\textbf{Explained variance in the linear and polynomial regression models.} Models are ordered by their mean predictive $R^2$ across objects. The explained variance represents the average performance on held-out data across all cross-validation folds. The bottom strip shows the per-class $R^2$ variance across models. Labels are displayed for the classes with the top-5 highest (red) and lowest (blue) $R^2$ variance  across models.}
  \label{figSI:r_squared_fits}
\end{figure}

\begin{figure}
  \centering
\begin{adjustbox}{max width=\textwidth}
  \includegraphics{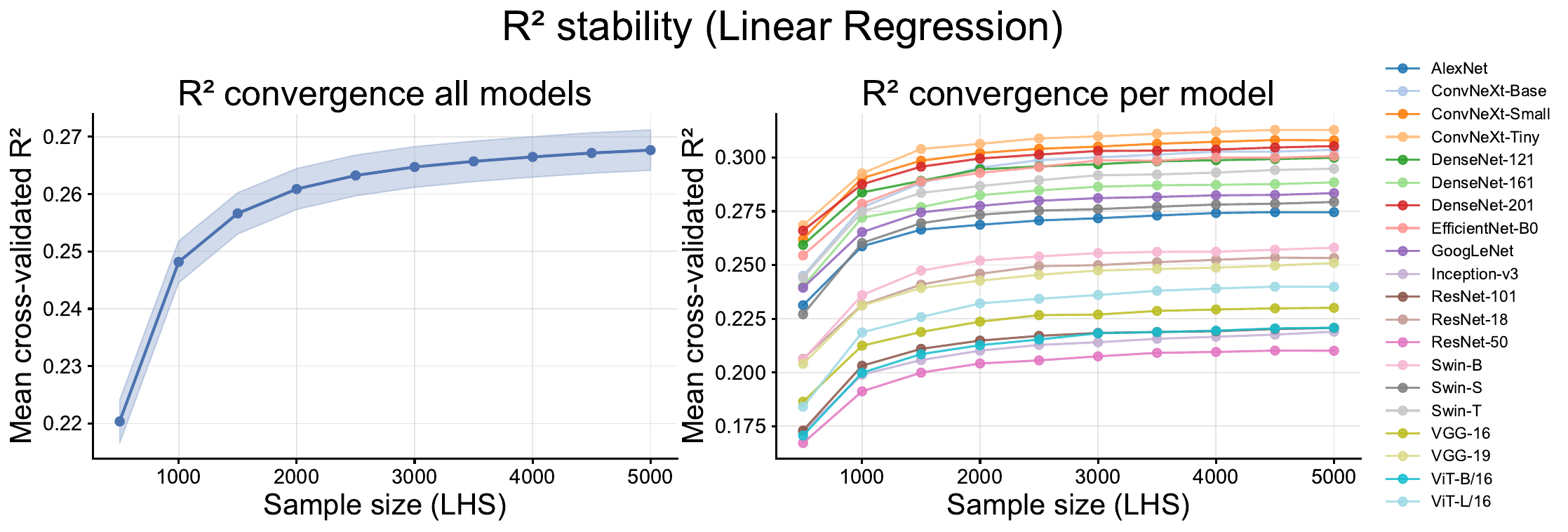}
  \end{adjustbox}
  \caption{\textbf{Convergence of goodness-of-fit ($R^2$) across sample size in linear regression.} For each model-mesh pair we fit a linear regression that predicts the model's decision margin from the nine scene parameters (full setup described in Section \ref{sec:evaluation}). We evaluate cross-validated $R^2$ using 10 folds as a function of the number of LHS samples used to fit the regression. \textbf{Left}: Mean cross-validated $R^2$ across all model-mesh pairs. The shaded band shows $\pm$ SEM across pairs. \textbf{Right}: Mean cross-validated $R^2$ per model, averaged across the 100 evaluated meshes.}
  \label{figSI:r2-convergence-linear}
\end{figure}

\begin{figure}
  \centering
\begin{adjustbox}{max width=\textwidth}
  \includegraphics{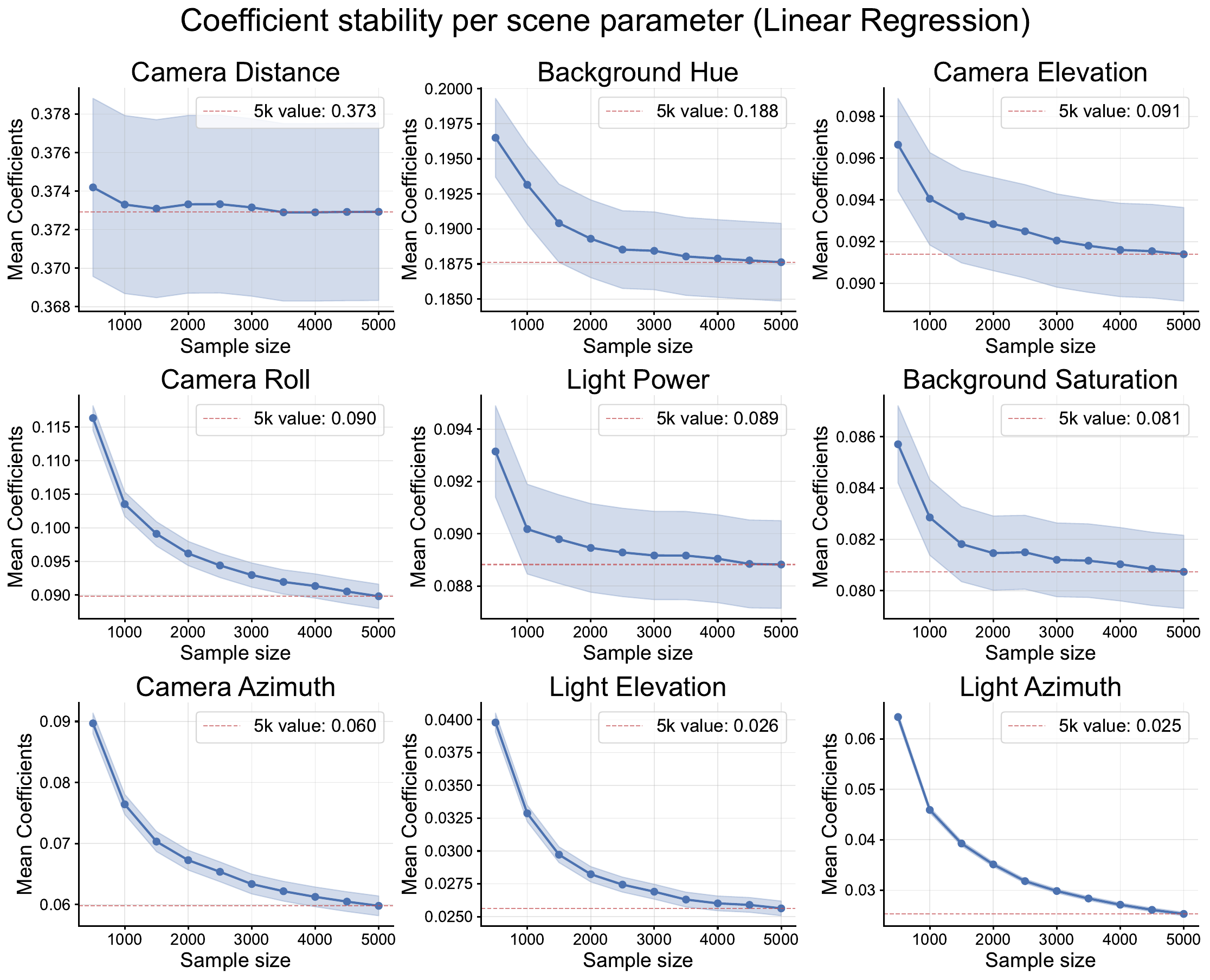}
  \end{adjustbox}
  \caption{\textbf{Coefficient stability of each parameter in linear sensitivity regression.} We compute the mean absolute regression coefficient for each parameter as a function of the LHS sample size used to fit the regression. Each subplot corresponds to one scene parameter. The shaded band indicates $\pm$SEM, and the red dashed line the value reached using 5000 samples.}
  \label{figSI:r2-convergence-params-linear}
\end{figure}

\begin{figure}
  \centering
\begin{adjustbox}{max width=0.8\textwidth}
  \includegraphics{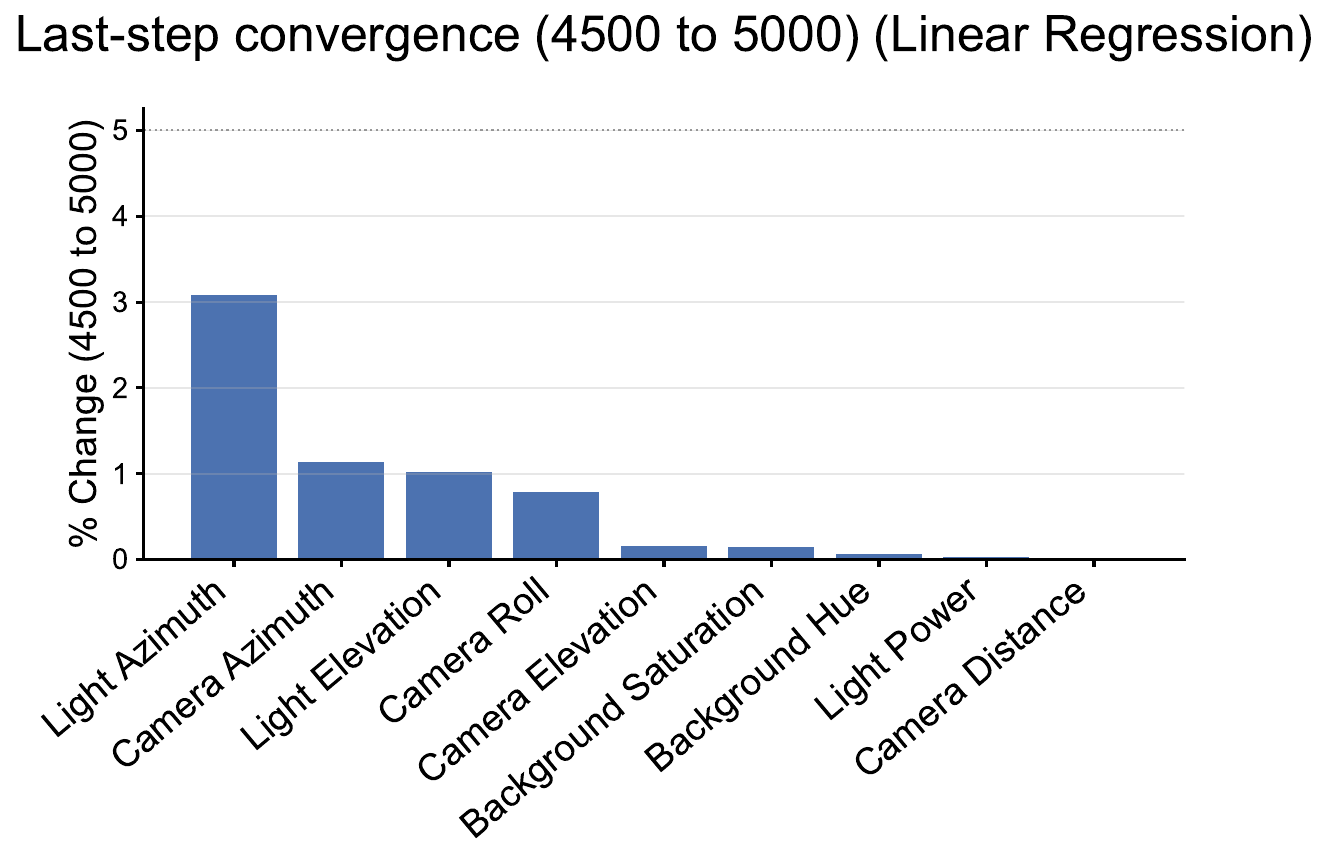}
  \end{adjustbox}
  \caption{\textbf{Convergence of regression coefficients from linear regression analyses.} For each scene factor, we show the relative change in mean absolute coefficient when increasing the number of Latin hypercube samples per mesh from 4500 to 5000, averaged across model-mesh pairs.}
  \label{figSI:stability-last-linear}
\end{figure}

\begin{figure}
  \centering
\begin{adjustbox}{max width=\textwidth}
  \includegraphics{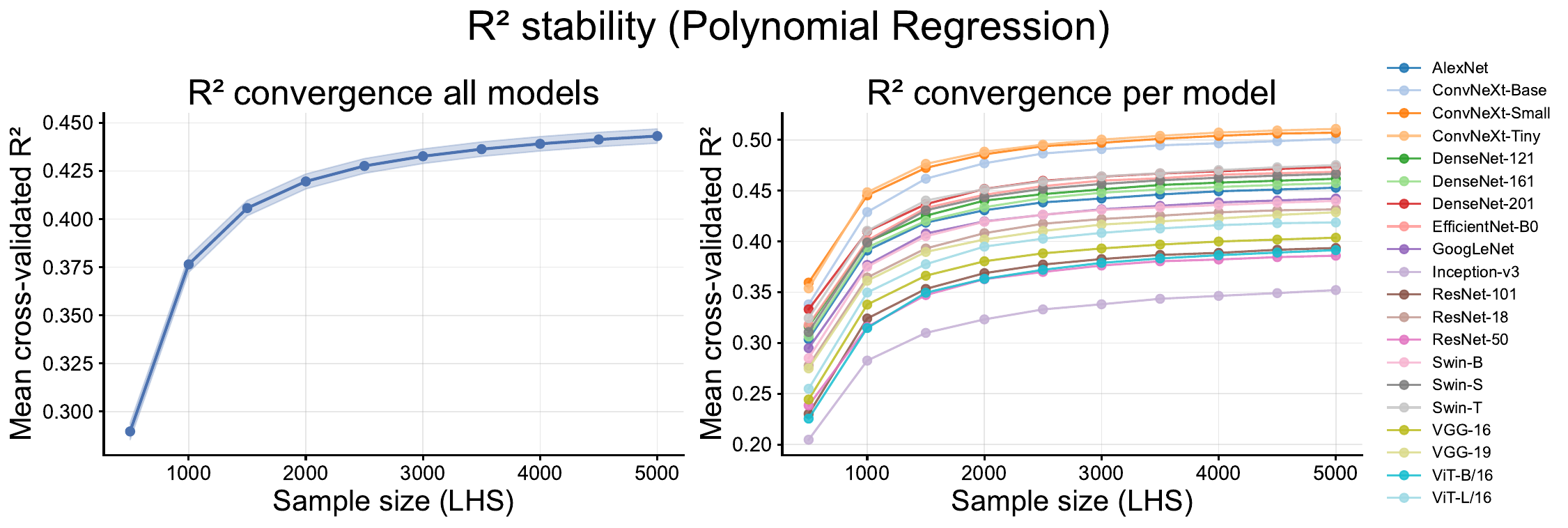}
  \end{adjustbox}
  \caption{\textbf{Convergence of goodness-of-fit ($R^2$) across sample size in polynomial regression.} As in Fig. \ref{figSI:r2-convergence-linear} but for polynomial regression.}
  \label{figSI:r2-convergence-poly}
\end{figure}

\begin{figure}
  \centering
\begin{adjustbox}{max width=\textwidth}
  \includegraphics{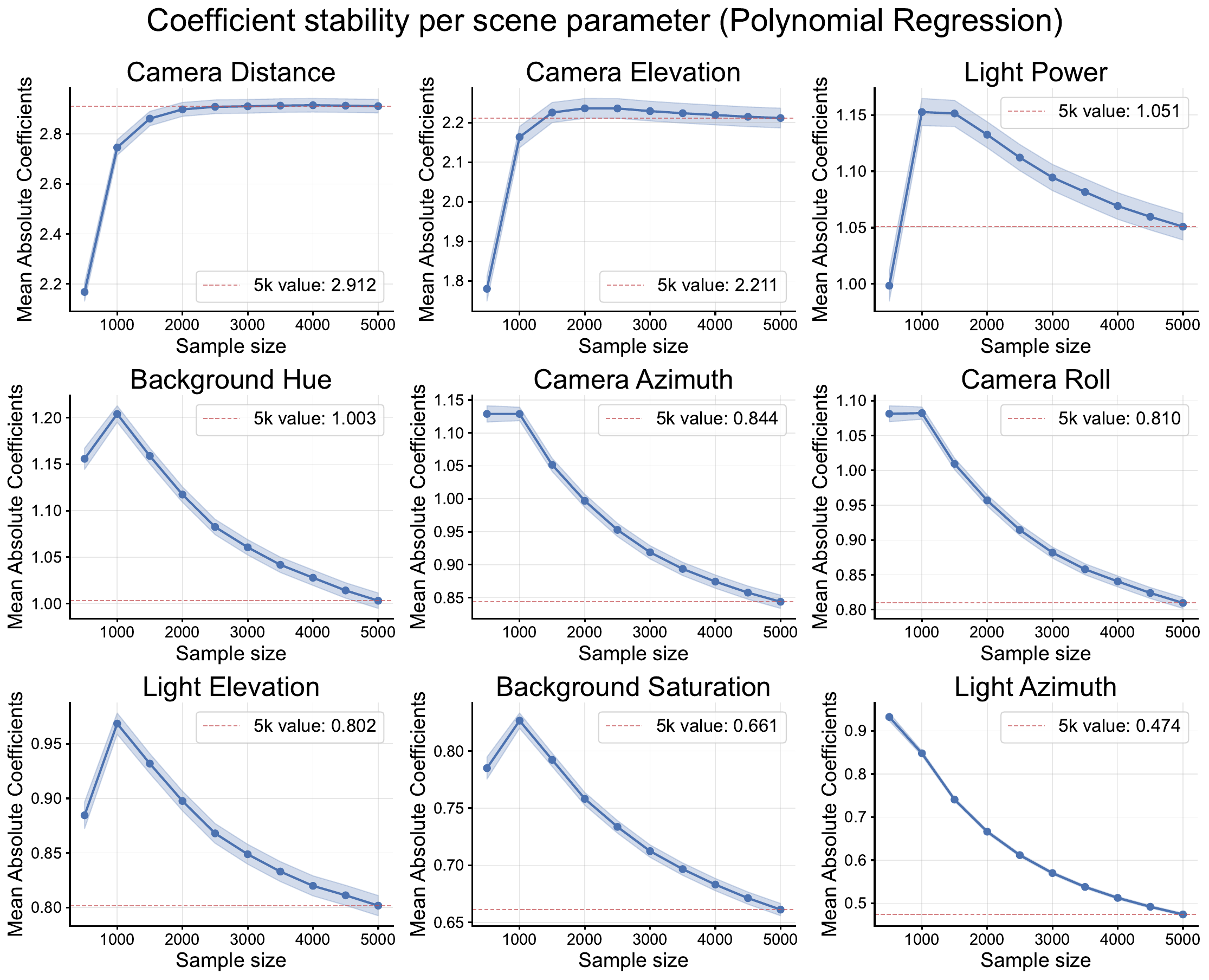}
  \end{adjustbox}
  \caption{\textbf{Coefficient stability of each parameter in polynomial sensitivity regression.} As in Fig. \ref{figSI:r2-convergence-params-linear} but for polynomial regression. }
  \label{figSI:r2-convergence-params-poly}
\end{figure}

\begin{figure}
  \centering
\begin{adjustbox}{max width=0.8\textwidth}
  \includegraphics{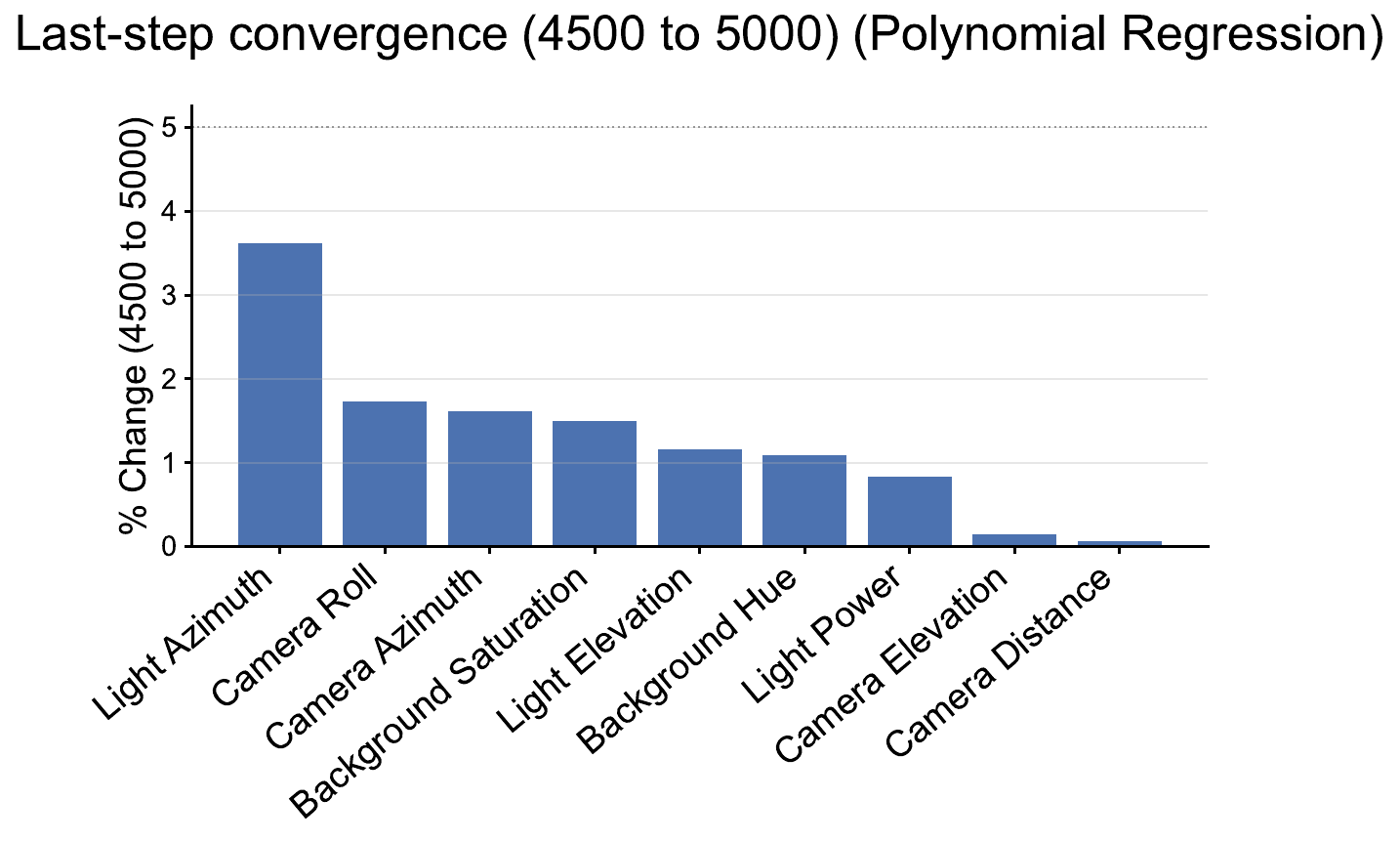}
  \end{adjustbox}
  \caption{\textbf{Convergence of regression coefficients from polynomial regression analyses.} As in Fig. \ref{figSI:stability-last-linear} but for polynomial regression.}
  \label{figSI:stability-last-poly}
\end{figure}

\begin{figure}
  \centering
\begin{adjustbox}{max width=\textwidth}
  \includegraphics{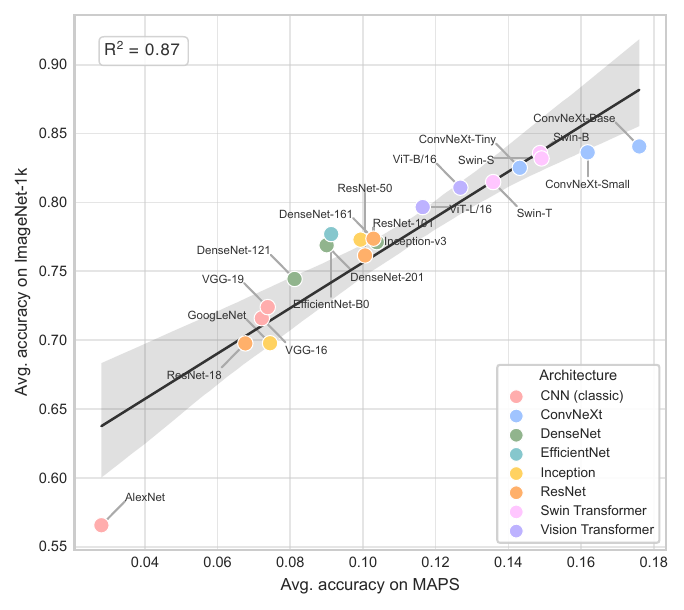}
  \end{adjustbox}
\caption{\textbf{Accuracy on MAPS vs. ImageNet-1k validation set.} Each point represents one of the evaluated models, with architecture family color-coded. Accuracy was averaged across the 100 evaluated classes for MAPS, and all classes for ImageNet-1k. Linear regression fit with 95\% confidence is shown.}
\label{figSI:acc-maps-vs-inet}
\end{figure}

\begin{figure}
  \centering
\begin{adjustbox}{max width=\textwidth}
  \includegraphics{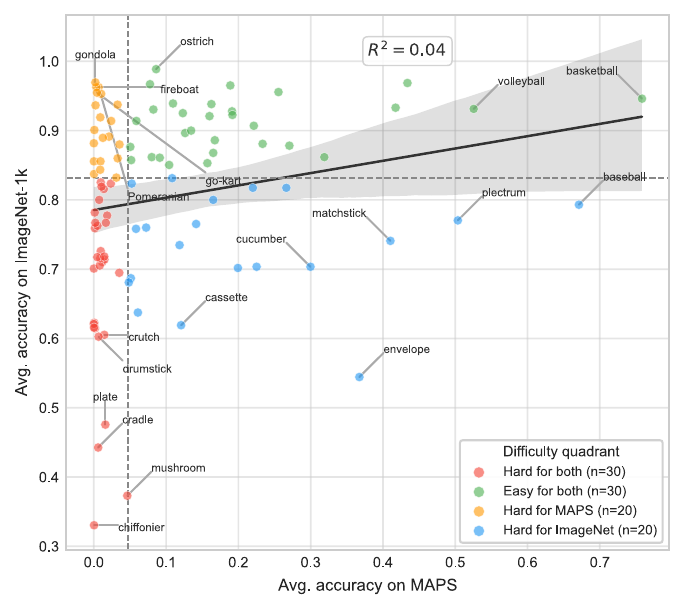}
  \end{adjustbox}
  \caption{\textbf{Per-class accuracy on MAPS vs. ImageNet-1k validation set.} Dashed lines correspond to the median accuracy for each dataset and partition the space into four difficulty quadrants.}
  \label{figSI:class-acc-maps-vs-inet}
\end{figure}

\clearpage

\end{document}